\documentclass[10pt,letterpaper]{article}
\usepackage[top=0.85in,left=2.75in,footskip=0.75in]{geometry}

\usepackage{amsmath,amssymb}
\usepackage{algorithm, algpseudocode}

\usepackage{changepage}

\usepackage{textcomp,marvosym}

\usepackage{cite}

\usepackage{nameref}
\usepackage[hidelinks]{hyperref}

\usepackage[right]{lineno}

\usepackage[nopatch=eqnum]{microtype}
\DisableLigatures[f]{encoding = *, family = * }

\usepackage[table]{xcolor}

\usepackage{array}

\newcolumntype{+}{!{\vrule width 2pt}}

\newlength\savedwidth

\raggedright
\usepackage[aboveskip=1pt,labelfont=bf,labelsep=period,justification=raggedright,singlelinecheck=off]{caption}

\makeatletter
\renewcommand{\@biblabel}[1]{\quad#1.}
\makeatother

\makeatletter
\newcommand{\multiline}[1]{%
  \begin{tabularx}{\dimexpr\linewidth-\ALG@thistlm}[t]{@{}X@{}}
    #1
  \end{tabularx}
}
\makeatother

\usepackage{lastpage,fancyhdr,graphicx,tabularx}
\usepackage{epstopdf}
\fancyheadoffset[L]{2.25in}
\fancyfootoffset[L]{2.25in}
\begin{document}
\vspace*{0.2in}

\begin{flushleft}
{\Large
\textbf\newline{Resolving uncertainty on the fly: Modeling adaptive driving behavior as active inference} % Please use "sentence case" for title and headings (capitalize only the first word in a title (or heading), the first word in a subtitle (or subheading), and any proper nouns).
}
\newline
\\
Johan Engstr{\"o}m\textsuperscript{1}*,
Ran Wei\textsuperscript{2},
Anthony D. McDonald\textsuperscript{3},
Alfredo Garcia\textsuperscript{2},
Matt O’Kelly\textsuperscript{1},
Leif Johnson\textsuperscript{1}

\bigskip
\textbf{1} Waymo LLC, Mountain View, CA, USA
\\
\textbf{2} Department of Industrial and Systems Engineering, Texas A\&M, College Station, TX, USA
\\
\textbf{3} Department of Industrial and Systems Engineering, University of Wisconsin-Madison, Madison, WI, USA
\\
\bigskip

% Insert additional author notes using the symbols described below. Insert symbol callouts after author names as necessary.
% 
% Remove or comment out the author notes below if they aren't used.
%
% Primary Equal Contribution Note


% Group/Consortium Author Note
%\textpilcrow Membership list can be found in the Acknowledgments section.

% Use the asterisk to denote corresponding authorship and provide email address in note below.
* jengstrom@waymo.com

\end{flushleft}
\section*{Abstract}
Understanding adaptive human driving behavior, in particular how drivers manage uncertainty, is of key importance for developing simulated human driver models that can be used in the evaluation and development of autonomous vehicles. However, existing traffic psychology models of adaptive driving behavior either lack computational rigor or only address specific scenarios and/or behavioral phenomena. While models developed in the fields of machine learning and robotics can effectively learn adaptive driving behavior from data, due to their black box nature, they offer little or no explanation of the mechanisms underlying the adaptive behavior. Thus, a generalizable, interpretable, computational model of adaptive human driving behavior is still lacking. This paper proposes such a model based on active inference, a behavioral modeling framework originating in computational neuroscience. The model offers a principled solution to how humans trade progress against caution through policy selection based on the single mandate to minimize expected free energy. This casts goal-seeking and information-seeking (uncertainty-resolving) behavior under a single objective function, allowing the model to seamlessly resolve uncertainty as a means to obtain its goals. We apply the model in two apparently disparate driving scenarios that require managing uncertainty, (1) driving past an occluding object and (2) visual time sharing between driving and a secondary task, and show how human-like adaptive driving behavior emerges from the single principle of expected free energy minimization.

% Please keep the Author Summary between 150 and 200 words
% Use first person. PLOS ONE authors please skip this step. 
% Author Summary not valid for PLOS ONE submissions.   
%\section*{Author summary}

%\linenumbers

% Use "Eq" instead of "Equation" for equation citations.
\section*{Introduction}
A fundamental aspect of motor vehicle driving, and locomotion in general, is to find an adequate balance between progress and caution. The main purpose of driving is typically to get to the destination as efficiently as possible. However, one also needs to make sure to get there without crashing and avoid other undesired consequences (e.g., getting ticketed). A key challenge here is to manage the uncertainty inherent in most traffic situations. For example, will the vehicle in front brake and, if so, how hard? Is there a risk of a pedestrian suddenly appearing behind the stopped bus ahead? Being too pessimistic about how a traffic situation may play out in uncertain situations could lead to overcautious behavior or even a complete lack of progress. Moreover, non-human-like, overcautious, behavior may be surprising to other road users with potential negative safety consequences \cite{dinparastdjadid2023measuring}. Yet, being too assertive in an uncertain situation may incur a significant risk of collision. 

Human drivers are most often able to manage this tradeoff, even in highly complex and uncertain traffic situations. Thus, understanding how human drivers manage uncertainty is critical for developing realistic and explainable computational models of their behavior. Such models have a range of applications such as explaining crash causation \cite{summala1996accident}, representing other road users in traffic simulation \cite{treiber2013microscopic, suo2021trafficsim, igl2018deep, markkula2023explaining}, establishing behavior benchmarks for autonomous vehicles \cite{engstrom2022modeling} or as part of the AV software itself \cite{sadigh2018planning}. 

Contemporary generative AI models can learn complex driving behaviors from large quantities of data (e.g., \cite{suo2021trafficsim, igl2022symphony}). There has also been extensive work in machine learning and robotics on developing models able to manage uncertainty through concepts like artificial curiosity and intrinsic motivation (e.g., \cite{schmidhuber1991curious, sun2011planning, hester2012intrinsically}). However, due to the black-box nature of these models, they do not lend themselves to explaining the cognitive mechanisms underlying human adaptive behavior, which is also typically not their purpose.

There is a long-standing tradition in traffic psychology in modeling adaptive road user behavior, such as the regulation of speed and headway. Models in this tradition, often referred to as motivational models, include the risk homeostasis theory \cite{wilde1982theory}, the zero-risk theory \cite{naatanen1976road, summala1988risk}, and the task capability interface (TCI) model \cite{fuller2005towards} (see \cite{lewis2012testing} for a review). In these models, excitatory forces such as the motivation to make progress towards the destination are balanced against inhibitory forces where uncertainty is typically a key component. However, these traditional motivational driving models are typically conceptual in nature, lacking mathematical rigor. 

One example of an early computational model of adaptive driving behavior, of particular relevance for this paper, is the pioneering work by Senders et al. \cite{senders1967distribution} based on the visual occlusion paradigm. In visual occlusion, subjects wear a helmet featuring a visor that, while driving, intermittently occludes the subject’s view. The occlusion and viewing times may be fixed, or the subject could be given control over the occlusion time by means of a switch that opens the visor for fixed viewing time period (typically 0.25-1 s). Senders et al. found that, when occlusion times were controlled at different levels, subjects adapted their speed such that shorter occlusion times resulted in higher speed and vice versa. Conversely, when subjects had voluntary control over the occlusion time, and speed was controlled, higher speeds led to shorter occlusion times and vice versa. To explain these results, the authors developed an information theoretic model based on the idea that uncertainty builds up during the occlusion intervals and that the observed adaptive behaviors (speed and occlusion interval adjustments) reflects the attempt of the driver to control this uncertainty. Senders et al. further proposed two main sources of  uncertainty in driving that need to be controlled: (1) the uncertainty about the traffic situation ahead due to loss of relevant visual information and (2) uncertainty about the vehicle’s position on the road due to random disturbances in vehicle lateral control. 

These early occlusion results have since been replicated for other types of driving scenarios. For example, in a recent study, Pekkanen et al. \cite{pekkanen2017task} found reduced voluntary occlusion times with reduced headway. Pekkanen et al. \cite{pekkanen2018computational} developed a computational model of drivers’ visual sampling where uncertainty about the consequences of action (acceleration), resulting from uncertainty in state estimation, is used to directly control attention (operationalized as a voluntary opening of the occlusion visor). 

Victor et al. \cite{victor2005sensitivity} analyzed drivers' visual time sharing between driving and a secondary task and found, in line with the occlusion studies, that the visual demand of driving during curve negotiation led to reduced off-road glance durations. Johnson et al. \cite{johnson2014predicting}, based on earlier computational modeling in non-driving domains \cite{sprague2003eye}, proposed a model of visual time sharing during car following based on a tradeoff between task priority and uncertainty. 

While most computational models of human adaptive driving behavior have focused on visual sampling, models with a more general behavioral scope are rare. One notable exception is the model by Kolekar et al. \cite{kolekar2020human}, based on zero-risk theory and the field of safe travel concept from Gibson and Crooks \cite{gibson1938theoretical}. The model represents uncertainty about potential collisions in terms of a dynamical risk field, and was demonstrated to account for a wide range of empirical adaptive driving behavior results reported in the literature. However, the current version of the model is limited to static scenarios with no other road users present.  

A common denominator in most of these existing models is that adaptive driving behavior is, on the one hand, driven by a motivation to achieve goals and, on the other, by the need to control uncertainty. The computational models reviewed above represent specific aspects of this phenomenon (e.g., visual sampling and time sharing). However, a generic computational model of adaptive driving behavior, applicable across all types of scenarios and behaviors, is still lacking.

In this paper, we propose such a model based on active inference, a behavior modeling framework originating from computational neuroscience \cite{friston2017active, parr2017uncertainty}. The application of active inference and the closely related predictive processing framework \cite{clark2015surfing, clark2013whatever, clark2023experience} in the automotive domain were explored in Engström et al. \cite{engstrom2018great}, but only on a conceptual level. In a series of recent papers \cite{wei2022modeling, wei2023world, wei2023active}, we have demonstrated how computational driver models based on active inference can be implemented and learned from data, thus offering a potential “middle ground” between traditional “black box” machine learning models and mechanistic human behavior models. In this paper, we focus specifically on how active inference can provide a conceptual and computational basis for modeling human adaptive driving behavior and, in particular, how a (Bayes) optimal balance between goal-directed action and the resolution of uncertainty emerges “automatically” from the minimization of expected free energy.

In active inference, the agent estimates the free energy associated with alternative future policies (defined as sequences of actions) within a defined planning time horizon and, at each time step, selects the action associated with the policy that has the lowest \textit{expected free energy} (EFE). Expected free energy can be formulated in several different ways (see \cite{friston2017active,friston2015active, parr2022active}. For present purposes we choose the formulation in Eq 1 which defines the expected free energy as the (negative) sum of a pragmatic value and an epistemic value, where the pragmatic value relates to goal seeking behavior and epistemic value to information-seeking (uncertainty-resolving) behavior, mapping conceptually to progress versus caution or exploitation vs. exploration.

\begin{align}\label{eq:efe}
    EFE = G(\pi) = -\underbrace{\mathbb{E}_{Q(\mathbf{o}|\pi)}[\log(P(\mathbf{o})]}_{\text{Pragmatic value}} - \underbrace{\mathbb{E}_{Q(\mathbf{s}, \mathbf{o}|\pi)}D_{KL}[Q(\mathbf{s}|\mathbf{o}, \pi) || Q(\mathbf{s}|\pi)]}_{\text{Epistemic value}}
\end{align}

In \eqref{eq:efe}, $\pi = a_{1:H}$ is a policy, $\mathbb{E}$ denotes expectations, $\mathbf{s} = s_{1:H}$ and $\mathbf{o} = o_{1:H}$ are sequences of future states and observations up to a lookahead time horizon $H$. $Q(\mathbf{s}|\pi)$ and $Q(\mathbf{o}, \mathbf{s}|\pi)$ are the ego’s belief about future state and state-observation sequences, respectively. The pragmatic and epistemic value terms in \ref{eq:efe} are further unpacked below. 

In the formulation of expected free energy in \ref{eq:efe}, the pragmatic value is defined based on a prior probability distribution over observations that is biased such that observations preferred by the agent have the highest probability and, hence, the highest pragmatic value. Thus, selecting policies that generate preferred observations will maximize the pragmatic value and contribute to minimizing the expected free energy. This hence implements a mechanism that generates goal-directed (or aversive) behavior, in a similar way as optimizing against a reward or cost function in optimal control or reinforcement learning \cite{sutton2018reinforcement}. 

The epistemic value represents the value of obtaining new information that may help to resolve uncertainty in the belief about future states. This may, in turn, enable (“open up”) new policies that maximize pragmatic value and thus realize the agent’s preferred observations. For example, when planning to overtake a car ahead, there is typically uncertainty about whether this will lead to a conflict with a potential vehicle approaching from behind in the adjacent lane. The uncertainty can be resolved by checking the rearview mirror to verify that the lane is clear. Epistemic value scores such information-seeking actions contributing to the overall expected free energy. In \ref{eq:efe}, the epistemic value of a policy is defined as the expected (Kullback-Leibler, KL) divergence ($D_{KL}$) between the agent’s prior and posterior beliefs about external states associated with that policy, corresponding to (expected) Bayesian Surprise \cite{itti2009bayesian, dinparastdjadid2023measuring}. Intuitively, this means that epistemic value is maximized for observations that lead to a maximum change in beliefs. Epistemic value can also be expressed as in (\ref{eq:epistemic}), as the difference between the posterior predictive entropy and the expected ambiguity (\cite{parr2022active}, p 135):
\begin{equation}
    \label{eq:epistemic}
    \mathbb{E}_{Q(\mathbf{o}|\pi)}D_{KL}[Q(\mathbf{s}|\mathbf{o},\pi)||Q(\mathbf{s}|\pi)]
    = \mathcal{H}[Q(\mathbf{o}|\pi)] - \mathbb{E}_{Q(\mathbf{s}|\pi)}\mathcal{H}[P(\mathbf{o}|\mathbf{s})]
\end{equation}
where $\mathcal{H}$ denotes Shannon entropy.

In (\ref{eq:epistemic}), the posterior predictive entropy (first term; $\mathcal{H}[Q(\mathbf{o}|\pi)]$) represents the uncertainty about future observations associated with a given policy. That is, a policy with a high posterior predictive entropy can lead to a variety of different observations and thus a strong potential to gain new information. The expected ambiguity (second term; $\mathbb{E}_{Q(\mathbf{s}|\pi)}\mathcal{H}[P(\mathbf{o}|\mathbf{s})]$) represents the expected diversity of observations for a given state. Intuitively, this means that the epistemic value of a policy is discounted if the state to be visited does not generate reliable observations (e.g., due to darkness or otherwise reduced visibility). Thus, epistemic value is maximized when the expected ambiguity is zero, that is, when the observation generated by the policy is expected to completely resolve the uncertainty.

Thus, in uncertain situations, minimizing expected free energy “automatically” promotes policies with high epistemic value, generating observations expected to resolve the uncertainty. As we will see below, a single action (e.g., moving forward) often carries both pragmatic (moving closer to the goal or away from danger) and epistemic value (getting a better view to resolve uncertainty). This leads to a key distinguishing feature of active inference: Goal directed (pragmatic) and information-seeking (epistemic) value are defined in a common currency and an optimal balance between them (given the agent’s beliefs and preferences) can be established by minimizing the expected free energy.

The key objective of this paper is to demonstrate how active inference can provide a novel conceptual and computational basis for modeling adaptive driving behavior. We explore how uncertainty can be resolved “on the fly” as an integral part of the general planning objective to minimize expected free energy. Specifically, we demonstrate how a model based on the single mandate to minimize expected free energy can account for two apparently disparate adaptive driving behaviors: (1) safely passing an occluding object and (2) visual time-sharing behavior, for example when texting on a cell phone.

\section*{Methods}
\subsection*{Overview}
\begin{figure}[!htb]
    \centering
    \includegraphics[width=1\textwidth]{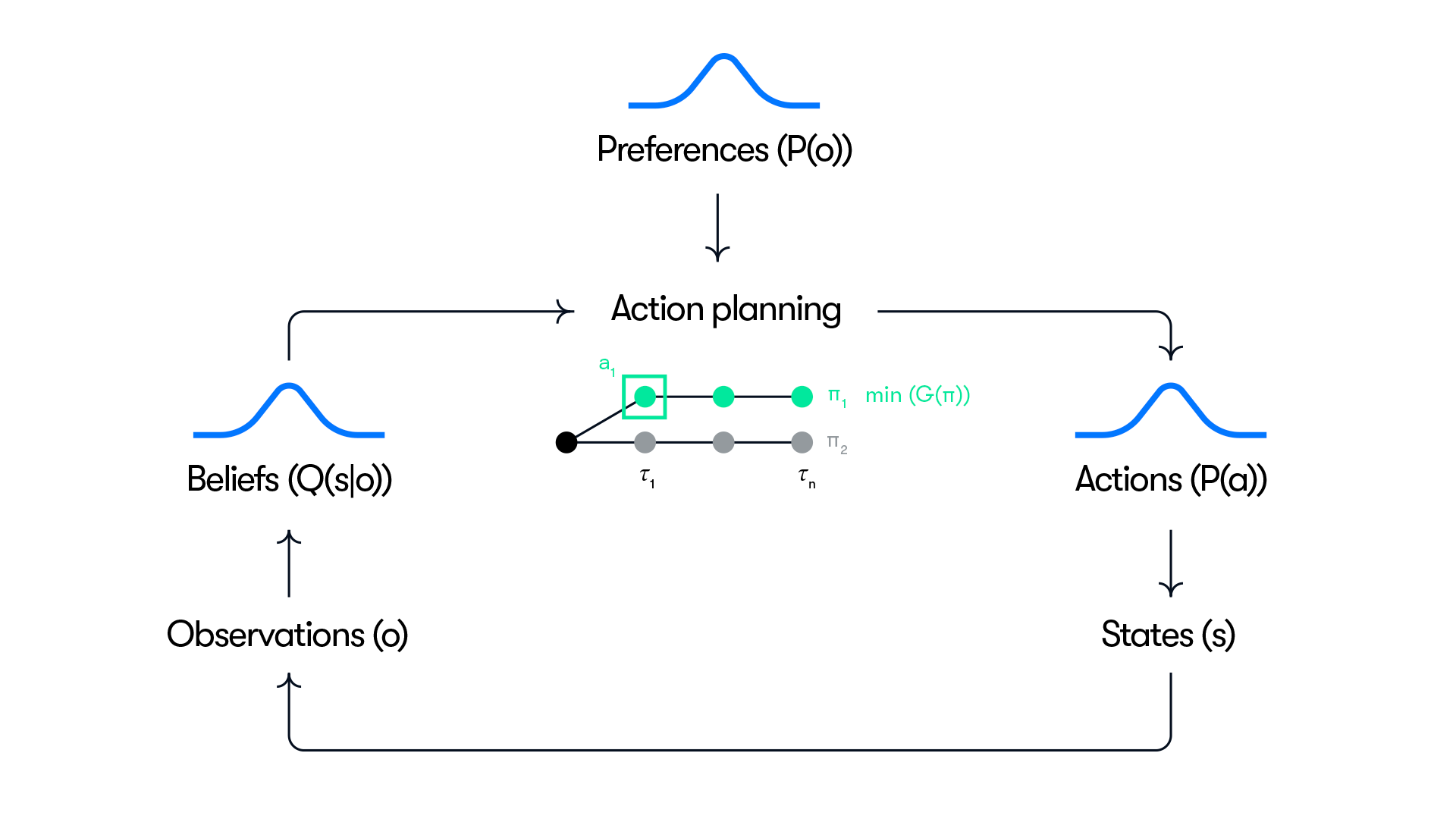}
    \caption{Schematic overview of the model. See the text for explanation.}
    \label{fig:framework}
\end{figure}

A conceptual overview of our model is given in Figure \ref{fig:framework}. Control actions (e.g. acceleration and steering inputs) are generated as the result of a planning process where policies ($\pi$), constituting sequences of future actions up to the planning horizon, are selected based on identifying the policy with the minimum expected free energy ($\min G(\pi)$). At each time step, the first action of the selected policy is executed. The action planning is based on the driver’s beliefs over hidden states ($Q(s)$) and preferences defined as priors over observations $P(o)$. The beliefs are continuously updated into posterior beliefs ($Q(s|o)$) based on new observations. The precision (inverse variance) of the beliefs represent the certainty of these beliefs over states. The preferences $P(o)$ define the observations that the driver is seeking to achieve through action (e.g., maintain a speed near the speed limit) and their precision represent the “strength”, or priority, of the preference (i.e., how motivated the driver is to keep the speed near the speed limit).

In action planning, a set of alternative counterfactual policies are sampled and rolled out up to a time horizon representing the planning window. In all simulations reported here, a planning horizon of 4 s is used. For each candidate policy, the beliefs over states are propagated forward in time from the current belief based on a state transition model and the propagated beliefs are used to calculate the counterfactual pragmatic and epistemic value at each future time step, $\tau$, in the planning window. The prior preferences are used to score the pragmatic value based on the counterfactual observations generated, where the pragmatic value at a given future time step, $\tau$, is computed based on the probability of the counterfactual observation made at that time step under the preference distribution. Similarly, epistemic value is computed based on counterfactual beliefs about future states and observations. The overall expected free energy of each policy is then scored based on (\ref{eq:efe}) and (\ref{eq:epistemic}) (as the negative sum of the epistemic and pragmatic values over the planning horizon), and the action at the next time step is sampled from a distribution of the first actions in the highest scoring policies. These general principles can be implemented in many different ways, where the current implementation is based on a particle filter, as described in more detail below.

The driver’s generative model is represented as a discrete-time Partially Observable Markov Decision Process (POMDP) which describes how hidden environment states (e.g., pose of the ego vehicle) evolve over time depending on the ego vehicle driver’s chosen policies and resulting control actions, and generate signals observed by the driver, for example the new pose of the ego vehicle or the presence of a pedestrian. Importantly, some of these state variables are not always directly observable to the driver. For example, the driver cannot observe a pedestrian when they are occluded by an object and cannot observe the road ahead while looking away from the road. 

The model uses a mixture of discrete and continuous state, observation and action variables with highly structured dependencies, which makes exact computation of the belief update and action selection intractable. We thus perform approximate belief update and policy/action selection using a particle filter and a particle planner. At a high level, this means that the model represents uncertainty about the hidden states using an ensemble of hypothetical states, where each ensemble member represents a different possibility. The model then selects the best actions by simulating future state-action-observation trajectories under different policies using a forward dynamics model and scores each policy for expected free energy using (\ref{eq:efe}) and (\ref{eq:epistemic}). Such a simulation-based inference method is known to approach the optimal solution with a large number of particles \cite{murphy2012machine}. 
 
\subsection*{Implementation}
In this section, we first describe the perception and action process of active inference agents. We then describe our particle-based implementation. 

\subsubsection*{Active Inference and Expected Free Energy}
We use the standard notation for POMDP \cite{kaelbling1998planning}, where $S=\{s\}$ denotes a set of states, $A=\{a\}$ denotes a set of actions, $O=\{o\}$ denotes a set of observations. The active inference agent has a generative model of the environment which consists of a state transition distribution $P(s'|s, a)$ and an observation distribution $P(o|s)$. 

Upon receiving an observation $o_t$, the agent updates its belief, defined as a probability distribution over the hidden environment state $Q(s_t)$, by minimizing the variational free energy of its generative model (see \cite{parr2022active}). The optimal belief distribution is known to have the following form \cite{da2020active}:
\begin{align}
    Q(s_t) \propto \exp\left(\log P(o_t|s_t) + \mathbb{E}_{Q(s_{t-1})}[\log P(s_t|s_{t-1}, a_{t-1})]\right)
\end{align}

Starting from the updated belief, the agent constructs predictions over future state-observation trajectories $(s_{t+1:t+H}, o_{t+1:t+H})$ for a lookahead horizon of $H$ time steps given a policy $\pi = a_{1:H}$. These predictions, defined over the lookahead time steps $\tau \in \{t+1, \dots, t+H\}$ in the form of probability distributions, can be constructed sequentially (i.e., via rollout) as follow:
\begin{align}\label{eq:rollout}
    Q(o_{\tau}, s_{\tau}|\pi) = \mathbb{E}_{Q(s_{\tau - 1})}[P(o_{\tau}|s_{\tau})P(s_{\tau}|s_{\tau-1}, a_{\tau-1})]
\end{align}

The quality of each policy is scored by the expected free energy function defined in (\ref{eq:efe}) as:
\begin{align*}%\label{eq:efe}
    G(\pi) = \sum_{\tau=1}^{H}-\underbrace{\mathbb{E}_{Q(o_{\tau}|\pi)}[\log(P(o_{\tau})]}_{\text{Pragmatic value}} - \underbrace{\mathbb{E}_{Q(s_{\tau}, o_{\tau}|\pi)}D_{KL}[Q(s_{\tau}|o_{\tau}, \pi) || Q(s_{\tau}|\pi)]}_{\text{Epistemic value}}
\end{align*}

\subsubsection*{Particle-Based Algorithm}
We use a particle-based approach to belief representation, inference, evaluation, and planning. We describe these components below and summarize the entire process as pseudocode in Algorithm 1.

\begin{algorithm}[!htb]
\caption{Simulation of particle-based active inference agent}\label{algo_btom}
\begin{algorithmic}[1]
\Require Environment model $\mathbb{P}(o|h, a)$,  agent transition model $P(s'|s, a)$, agent observation model $P(o|s)$, agent preference model $P(o)$, number of belief particles $N$, number of planning particles $\tilde{N}$, number of plans to evaluate $M$, percent of top plans to retain $x$, number of planning iterations $K$. 

\State Initialize environment observation $o_0$
\State Initialize belief particles $b_{0} = (\delta_{0, 1:N}, w_{0, 1:N})$
\For{$t=0:T$}
    \State \textcolor{blue}{\# Agent planning}
    \State Initialize planning particles $\tilde{\delta}_{1:\Tilde{N}}$ using multinomial resampling from $b_{t}$
    \For{$k=1:K$} %\Comment{CEM planning}
        \State Sample $M$ action sequences $\{a_{m, 1:H}\}_{m=1}^{M}$ from distribution $P(a_{1:H})$
        \For{$m=1:M$} %\Comment{\textcolor{blue}{Evaluate plan}}
            \State \textcolor{blue}{\# Evaluate plan}
            \State \multiline{%
            Sample a sequence of state and observation particles $(\tilde{\delta}_{1:\tilde{N}, 1:H}, \tilde{o}_{1:\tilde{N}, 1:H})$ using forward rollouts of $P(s'|s, a)$ and $P(o|s)$}
            \State Compute the EFE of $(\tilde{\delta}_{1:\tilde{N}, 1:H}, \tilde{o}_{1:\tilde{N}, 1:H})$
        \State \multiline{%
        Refit distribution $P(a_{1:H})$ to the top $x$ percent of the $M$ action sequences by EFE}
        \EndFor
    \EndFor
    \State \textcolor{blue}{\# Choose action and update environment}
    \State Choose $a_t$ as the mean of $P(a_0)$
    \State Sample observation $o_{t+1} \sim \mathbb{P}(\cdot|h_t, a_t)$
    \State \textcolor{blue}{\# Update belief using particle filter}
    \State Propagate belief particles forward: $\delta_{t+1, n} \sim P(\cdot|\delta_{t, n}, a_t)$
    \State Update weights: $w_{t+1, n} \propto w_{t, n} P(o_{t+1}|\delta_{t+1, n})$
    \State Apply systematic resampling on particles and weights if $N/(1 + \sum_n w_n^2) \leq N/2$
\EndFor
\end{algorithmic}
\end{algorithm}

Using a particle filter, we represent the ego’s belief $b_t$ over all state variables using a set of $N$ particles $\{\delta_1, ..., \delta_N\}, \delta_n \in R^{|S|}$, where each particle consists of a vector corresponding to a realization of the state variables and a weight $w_n \geq 0, \sum_{n} w_n = 1$ representing the likeliness of the realization under the ego’s belief distribution. Under this representation, the ego belief has high certainty (high precision) if all particles with high weights are similar, e.g., all particles correspond to the pedestrian being present, and low certainty if all particles have even weights and are substantially different, e.g., the element representing the pedestrian’s position in each particle is evenly distributed on the road map. 

Upon executing an action and receiving a new observation $o_{t+1}$, we update the set of particles, including both the state vectors and weights, using a Sequential Importance Resampling (SIR) filter \cite{murphy2012machine}. The SIR filter first samples a next state conditioned on the current state and action and updates the weights using the following equation:
\begin{align}
    w_{t+1, n} \propto w_{t, n} P(o_{t+1}|\delta_{t+1, n})
\end{align}
where $\delta_{t+1, n} \sim \mathbb{P}(\cdot|\delta_{t, n}, a)$. 

Then, it resamples the current set of particles if the effective sample size: $N_\text{eff} \approx N/(1 + \sum{w^2})$ is less than N/2. To address the mode collapse problem common in particle filters, we use a large number of samples and apply systematic resampling so that the sampled particle weights are equidistant. In this way, low weight particles are still represented.

\subsubsection*{EFE computation}
Evaluating the EFE of an action sequence under a particular belief requires propagating the belief particles to compute the state and observation distributions at each counterfactual future time step and using the propagated particles to evaluate the pragmatic and epistemic values. 

The particles can be easily propagated by recursively sampling the next state from the transition distribution conditioned on the previous sampled state and policy action and then sampling the next observation from the observation distribution conditioned on the sampled next state, i.e., (\ref{eq:rollout}). To evaluate the pragmatic value (the first term 1 in (\ref{eq:efe})), we compute the average log likelihood of each observation sample under the preference distribution. 

To compute the epistemic value (term 2 in (\ref{eq:efe})), we use the decomposition of the expected information gain in (\ref{eq:epistemic}),  which is the difference between the posterior predictive entropy and the expected ambiguity: 
\begin{align*}
    \mathbb{E}_{Q(o|\pi)}D_{KL}[Q(s|o,\pi)||Q(s|\pi)]
    = \mathcal{H}[Q(o|\pi)] - \mathbb{E}_{Q(s|\pi)}\mathcal{H}[P(o|s)]
\end{align*} 
We approximate the intractable posterior predictive entropy using a Kernel Density Estimator (similar to \cite{fischer2020information}).

\subsubsection*{Model predictive control}
Given each updated belief, we compute the approximately optimal EFE-minimizing action using the Cross Entropy Method (CEM) for model predictive control \cite{de2005tutorial}. CEM iteratively refines a distribution over action sequences, i.e., policies, by sampling a batch of action sequences, simulating their trajectories forward, selecting the top $x$ percent scoring trajectories, and refitting the action distribution to the selected action sequences. This process can be understood as sampling from a distribution of action sequences in proportion to their EFE values similar to relative entropy policy search \cite{peters2010relative}.

Our application of CEM has two major differences from its normal use in optimal control and trajectory optimization. First, the optimal decision for an active inference agent is based on its belief as opposed to a known state. Thus, we adapt the default CEM by treating a set of belief particles as the state. Second, we use a mixture of discrete (gaze direction) and continuous actions (vehicle control) in the visual time sharing scenario, whereas the default CEM typically only optimizes continuous actions. We solve this by separately fitting the discrete and continuous variables once the best trajectory samples are selected in each iteration. 

\section*{Results}
This section describes the simulation results from applying our model to two different driving scenarios that require the control of uncertainty: (1) passing an occluding object and (2) visually time sharing gaze between driving and a secondary task. In the first scenario, the uncertainty concerns the potential presence of a pedestrian hidden behind the object who may step into the ego vehicle’s path. In the second scenario, uncertainty about the lateral position of the vehicle in the lane builds up during glances away from the road due to disturbances such as wind gusts and an uneven road surface \cite{senders1967distribution}.

\subsection*{Scenario 1: Passing an occluding object}
\begin{figure}[!htb]
    \centering
    \includegraphics[width=1\textwidth]{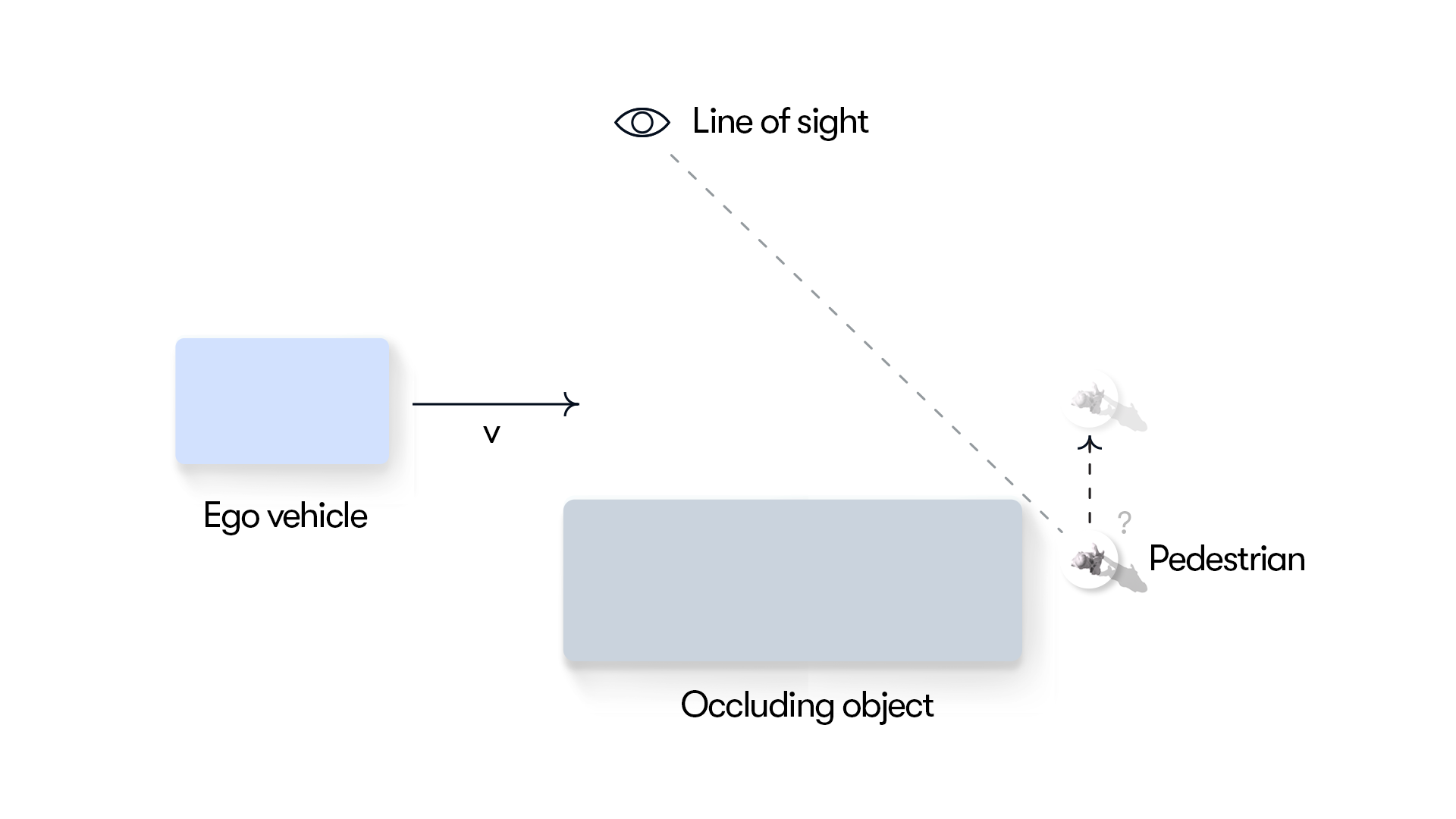}

    \includegraphics[width=1\textwidth]{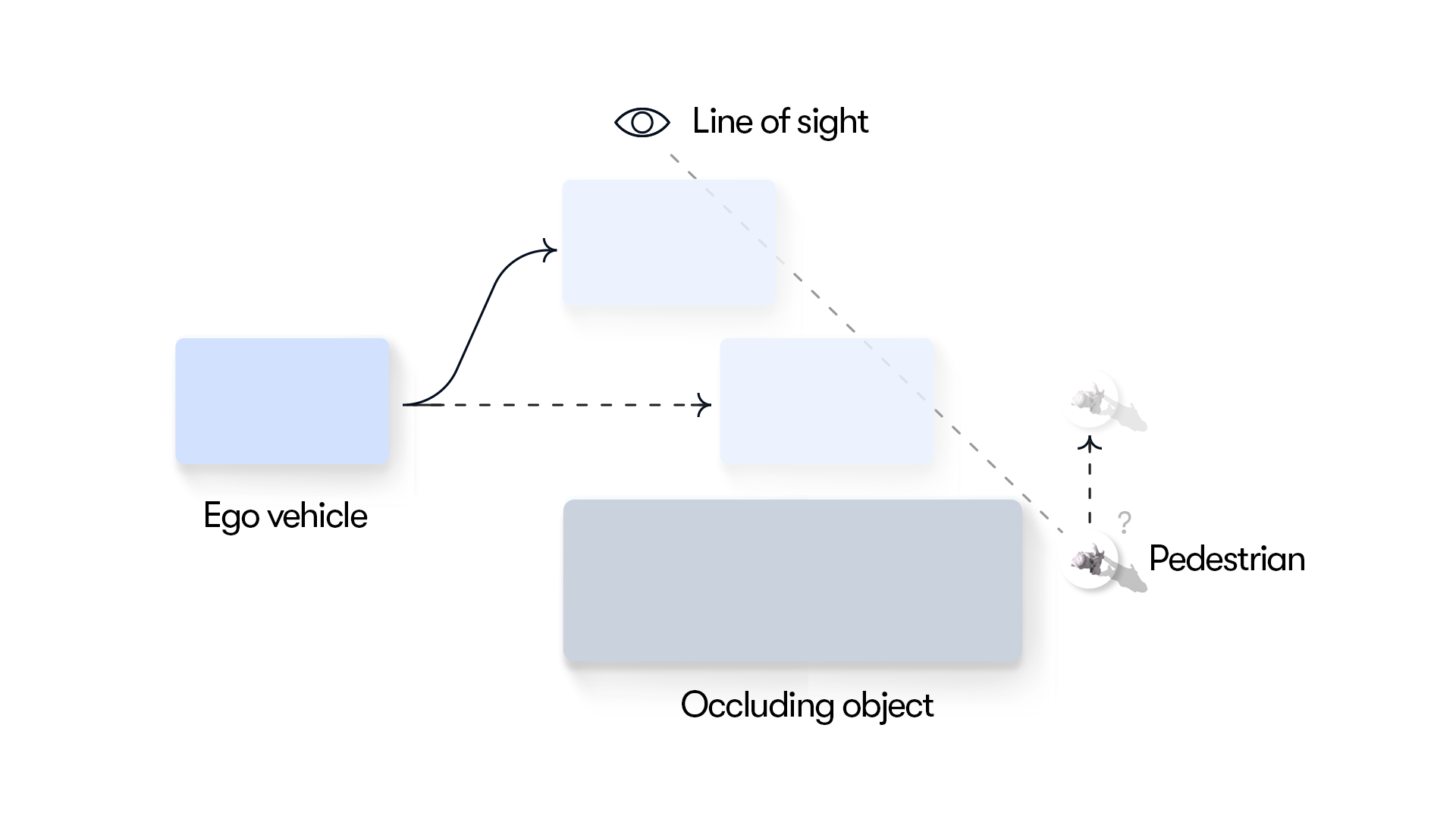}
    \caption{Conceptual illustration of Scenario 1. \textbf{Top:} The ego vehicle driver ideally wants to keep speed as close as possible to the speed limit. However, until reaching the line of sight, the driver is uncertain whether a pedestrian is hidden behind the double-parked vehicle and thus needs to adapt their speed to make sure they can stop short of the location where a pedestrian may appear. \textbf{Bottom:} By moving left the ego driver could reach the line of sight earlier, thus resolving the uncertainty sooner and potentially enabling more efficient passing of the occluding object.}
    \label{fig:line_of_sight}
\end{figure}

In this scenario, the ego vehicle approaches a large occluding object (e.g., a parked bus) and there is uncertainty about whether a pedestrian, potentially hidden behind the object, will encroach into the ego vehicle’s path (see Figure \ref{fig:line_of_sight}). We make the simplifying assumptions that a single pedestrian is the only possible obstacle that could be hidden behind the object and that, if a pedestrian is present, it will always step out in the road and cause a potential conflict with the ego vehicle. We further assume that the pedestrian can only be present at a given point along the horizontal (x) dimension so that it, if present, always becomes visible when it gets into the line of sight of the ego vehicle driver (see Figure \ref{fig:line_of_sight}). These assumptions simplify the current model implementation but they do not impose any fundamental limitations on the general modelling framework.

We further assume that, as a default, the ego vehicle driver prefers to keep the speed at the speed limit to maximize progress while respecting the rules of the road. However, if the driver believes that there is a risk for a hidden pedestrian encroaching into their path, they need to adapt their speed to be able to stop well ahead of the pedestrian (to meet the preference of conflict avoidance) without harsh braking (to meet their preferences for deceleration comfort). When the ego vehicle reaches the line of sight, the driver's uncertainty about the presence of the pedestrian is resolved and, if no pedestrian is present, they can speed up again to the preferred speed. Furthermore, as shown in Figure \ref{fig:line_of_sight} (bottom), by moving left in its lane the ego driver can reach the line of sight earlier, thus resolving the uncertainty that hinders progress and enabling a potentially faster trajectory past the occluding object. The general goal of the current simulation is to demonstrate how these adaptive driving behaviors emerge solely on the basis of minimizing expected free energy.

\subsubsection*{Model specification} %was model setup proposed change
The state, observation and action variables in the model are listed in Table \ref{tab:table1}. The ego vehicle kinematic states are denoted as $s^{ego} = [x, v_x, a_x, y, v_y, a_y]$ and are assumed to follow linear dynamics. The pedestrian’s position $s^{ped}=[x, y]$ is determined by a context variable, $I$, denoting whether the pedestrian is present or not present. If the pedestrian is present, its x and y positions will be set next to the occluding object. Otherwise, its x and y position will be set to a null value far away from the ego vehicle (e.g., -1000). $C$ represents whether there is a conflict between the ego and the pedestrian or if the vehicle exits the lane, in which case $C$ is set to 1, and 0 otherwise. A conflict is defined as the pedestrian being present and the longitudinal distance between the ego and the pedestrian is less than a safe distance (2 m). 

As described above, the main source of uncertainty in this scenario is that the presence or absence of the pedestrian cannot be determined before the vehicle reaches the line of sight. Geometrically, as shown in Figure \ref{fig:line_of_sight}, the pedestrian is occluded if it is behind (i.e, with smaller x coordinate) the line of sight connecting the ego vehicle and the upper right tip of the occluding object (since, as described below, we are modeling the ego vehicle as a point mass which is also the reference point from which the observation is made). For the observation space, we introduce a variable $o^{I}$ to represent the context observed by the ego. $o^{I}$ is set to “observed” when the ego vehicle enters the line of sight, otherwise, it is set to “not observed”. The rest of the observation variables share the same semantics and dimensionality as the underlying state variables, except when the pedestrian is occluded. In this case, the observations of the pedestrian's position, $o^{ped}$, are set to null. Hence, as long as there is no occlusion, we assume no uncertainty over the observations and that, hence, the ego can observe both its own and the pedestrian’s kinematics states exactly. 

\begin{table}[ht]
\caption{State, observation and action variables in the POMDP for Scenario 1.}
\label{tab:table1}
\resizebox{\textwidth}{!}{\begin{tabular}{lcll}
\hline
Variable name & Symbol & Values & Type \\ \hline
\multicolumn{4}{l}{States} \\ 
\hline
Pedestrian context & $I$ & not present (0), present (1) & Discrete \\
Is conflict or lane exit & $C$ & yes (1), no (0) & Discrete \\
Pedestrian position & $s^{ped}$ & xy position & Continuous \\
Ego kinematics & $s^{ego}$ & xy position, xy speed, xy acceleration & Continuous \\ 
\hline
\multicolumn{4}{l}{Observations} \\ \hline
Pedestrian context observation & $o^{I}$ & Not observed (1), observed (2) & Discrete \\
Conflict observation & $o^{C}$ & yes (1), no (0) & Discrete \\
Pedestrian position observation & $o^{ped}$ & xy position, null & Continuous \\
Ego kinematics observation & $o^{ego}$ & xy position, xy speed, xy acceleration & Continuous \\ 
\hline
\multicolumn{4}{l}{Action} \\ \hline
Ego control & $a^{ego}$ & xy acceleration & Continuous \\ 
\hline
\end{tabular}}
\end{table}

As described above, the model’s preferences are defined as priors over observations. The preferences of the model used in Scenario 1 and their default values are given in Table \ref{tab:Table2}. All simulations were run with a 200 ms time step (i.e, an update frequency of 5 Hz). Due to the stochasticity of the model, each simulation run yields a unique trace. However, for clarity, we only plot randomly selected single simulation traces. 

\begin{table}[htb]
\caption{Preference priors for Scenario 1.}
\label{tab:Table2}
\resizebox{\textwidth}{!}{\begin{tabular}{llc}
\hline
Preference prior & Specification & Default values \\ 
\hline
Speed keeping & \begin{tabular}[c]{@{}l@{}}Gaussian distribution \\ centered at the speed limit\end{tabular} & \begin{tabular}[c]{@{}l@{}}$\mu=10 m/s$\\ $\sigma=1 m/s$\end{tabular} \\
Lane keeping & \begin{tabular}[c]{@{}l@{}}Triangular distribution \\ centered at 0 and bounded at the lane boundaries\end{tabular} & N/A \\
Acceleration & \begin{tabular}[c]{@{}l@{}}Gaussian distribution \\ centered at zero for accelerations (x \& y)\end{tabular} & \begin{tabular}[c]{@{}l@{}}$\mu=0 m/s^2$\\ $\sigma=0.5 m/s^2$\end{tabular} \\
Conflict & \begin{tabular}[c]{@{}l@{}}Categorical distribution \\ representing an absolute preference over “no-conflict”\end{tabular} & N/A \\ 
\hline
\end{tabular}}
\end{table}

\subsubsection*{Simulations}
This section presents the results of different permutations of the occlusion scenario with the purpose to illustrate the key aspects of our model described above. We begin with the case where the ego driver can only drive straight (i.e., not move laterally, e.g., due to a narrow lane) and then extend this to allow for lateral movement that enables the driver to resolve uncertainty earlier through epistemic action (moving left to reach the line of sight earlier). In all simulations, we assume that no pedestrian is {\it actually} present while the ego driver model may, or may not, initially {\it believe} that a pedestrian could be present with a given probability.
\paragraph{Simulation 1a: Safely passing an occluding object.}
The purpose of this initial simulation is to show how our active inference model generates successful adaptive behavior in the occlusion scenario, finding an optimal balance between progress and caution given its set beliefs and preferences. Since, in this first simulation, the driver can only drive straight and not move laterally, the ego driver’s preferences reduce to prior distributions on preferred speed, comfortable accelerations and avoiding conflicts. 

\begin{figure}[!htb]
    \centering
    \includegraphics[width=1\textwidth]{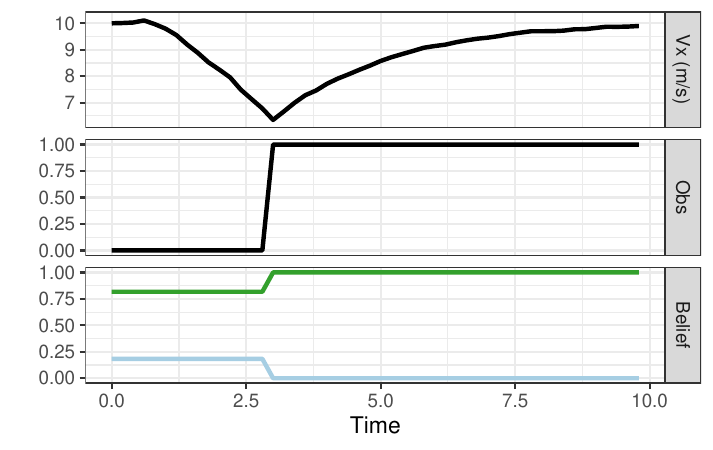}
    \caption{Results from simulation 1a with a model that initially believes there is a 20\% chance of a pedestrian being present behind the occluding object. Bottom panel: Evolution of the model’s beliefs about the pedestrian presence (blue: pedestrian present; green: pedestrian absent). Middle panel: Observation of the area behind the occluding object. Top panel: Vehicle speed selected by the model. See the text for explanation.}
    \label{fig:1a}
\end{figure}

We initiate the model at the speed limit (10 m/s) and a 20\% belief that the pedestrian is present. We assume that this belief matches the true probability that a pedestrian is present (in other words, the generative model matches the generative process). As shown in Figure \ref{fig:1a} (top panel), the model initially slows down and then speeds up again after the uncertainty about the presence of the pedestrian has been resolved (bottom panel) when reaching the line of sight and it can be observed that no pedestrian is present (middle panel). 

In our model, the initial slowing down behavior results from selecting a policy that maximizes the probability of preferred observations (maintaining the preferred speed, avoiding harsh decelerations and avoiding conflicts) taking into account the initial uncertainty in the belief about the presence of the pedestrian. Uncertainty about the presence of the pedestrian is represented by the dispersion of belief particles on the pedestrian context and position variables, and these variables are not updated by the particle filter until the ego vehicle reaches the line of sight. Once the uncertainty about the presence of the pedestrian has been resolved (when the ego reaches the line of sight) the policies with the highest pragmatic value are those that generate speed observations close to the preferred speed (i.e., the 10 m/s speed limit) and thus the model speeds up again. In this case, the behavior (policy selection) is driven solely by maximizing pragmatic value, thus minimizing expected free energy (\ref{eq:efe}). The model’s behavior (moving forward) also carries epistemic value as it eventually brings the pedestrian site within its line of sight, but this epistemic value here “comes for free” with the move-forward policy that maximizes the pragmatic value. Thus, in this scenario, pragmatic value does not have to be traded against epistemic value. This illustrates the key point that a given behavior often carries both pragmatic and epistemic value. The simulation in Figure \ref{fig:1a} provides a simple first illustration for how minimizing expected free energy (in this case by maximizing expected pragmatic value) leads to adaptive behavior that is optimal {\textit given the ego driver’s preferences and beliefs}.

\begin{figure}[!htb]
    \centering
    \includegraphics[width=1\textwidth]{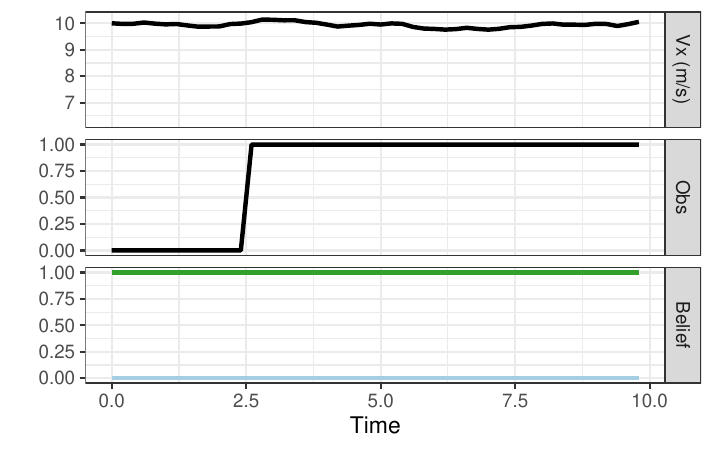}
    \caption{Results for simulation 1b, for a model with false certainty that no pedestrian is present. Otherwise similar to Figure \ref{fig:1a} (see text for explanation).}
    \label{fig:1b}
\end{figure}

\paragraph{Simulation 1b: False certainty}
We now change the ego driver model’s prior belief from from a 20\% to a 0\% belief that a pedestrian is present. Since we assume that there is a true 20\% chance that a pedestrian may appear, the model now has a false (overly certain) belief to the contrary. In other words, this represents a mismatch between the prior belief and the true statistics of the situation (i.e., the generative model does not match the generative process). As shown in Figure \ref{fig:1b}, as a result of the false belief (bottom panel) the ego maintains a constant speed near the speed limit (top panel) as it drives through the line of sight thus ignoring the potential presence of the pedestrian (middle panel).

In terms of our model, due to the lack of uncertainty in the beliefs about the pedestrian (and the resulting false certainty about the lack of risk for a conflict), the policy that minimizes the expected free energy is that which maximizes progress, in this case maintaining the preferred speed past the occlusion. This can be seen as an example of “optimal behavior with a sub-optimal model” \cite{schwartenbeck2016inference}. Behavior is optimal given the ego driver’s beliefs, but the beliefs do not match the true statistics of the environment, that is, the ego driver has a false certainty about the absence of pedestrians which leads to overly assertive behavior. Real world examples of this phenomenon include novice drivers that have not yet learned the true statistics (generative process) of all traffic situations or a drunk driver with a biased generative model (generating overly certain beliefs), resulting in reduced risk aversion.
\paragraph{Results for simulation 1c: Harvesting epistemic value through lateral movement
}
In Simulations 1a and 1b, policy and action selection was driven by pragmatic value only. However, in many real world situations, actions differ in their pragmatic and epistemic value and the optimal policy may, for example, involve an initial action yielding mainly epistemic value (e.g., turning on the light in a dark room, checking the rear view mirror) which resolves the uncertainty and enables pragmatic actions that realize the agent’s goal (exit the room, overtake the vehicle ahead). To illustrate how such policies, where uncertainty is resolved “on the fly”, can be found by minimizing expected free energy, we introduce the possibility for the ego vehicle to move laterally (along the y-axis). Due to the geometry of the situation (see Figure \ref{fig:line_of_sight}), the lateral movement allows the ego driver to resolve the uncertainty about the potential pedestrian earlier (by reaching the line of sight earlier; see Figure \ref{fig:line_of_sight}, bottom). This may lead to a more efficient overall path past the occluding object; even if the initial lateral movement temporarily reduces the pragmatic value (due to the agent’s prior preference to stay in the middle of the lane) this may be offset by the advantage of being able to speed up earlier (thus satisfying the preference of maintaining a speed close to the speed limit).

The driver’s preferences are the same as in 1a and 1b with the addition of a lane keeping preference defined by a triangular distribution centered at 0 and bounded at the lane edges, with a lane width of 3 m. The driver’s prior belief is the same as in Simulation 1a, that is, the driver believes with 20\% probability that a pedestrian would appear.

\begin{figure}[!htb]
    \centering
    \includegraphics[width=1\textwidth]{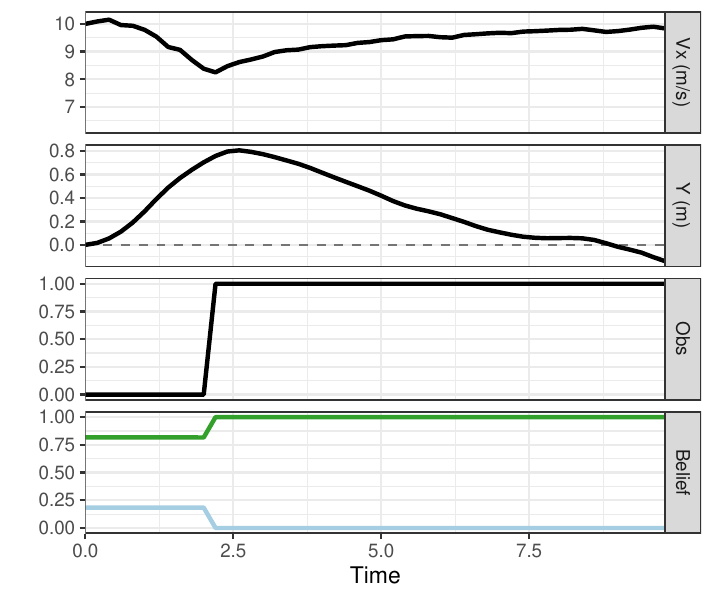}
    \caption{Simulation of epistemic action in 1c. The vehicle lateral position is added but otherwise the same as Figure 3 and 4.}
    \label{fig:1c}
\end{figure}

The simulation results are shown in Figure \ref{fig:1c}. The model initially slows down as in Simulation 1a (top panel) but also moves to the left, reaching the line of sight earlier. This allows the model to resolve the uncertainty one second earlier than in Simulation 1a (around 2 s compared to 3 s in Simulation 1 a; see the bottom panels of Figure \ref{fig:1a} and Figure \ref{fig:1c}) and is thus also able to speed up earlier than in Simulation 1a.

In our model, the leftward movement is generated by the possibility of the driver to harvest epistemic value, which contributes to minimizing the expected free energy (\ref{eq:efe}). As described above (\ref{eq:epistemic}), the epistemic value can be decomposed as the difference between the posterior predictive entropy and the expected ambiguity \cite{parr2022active}. In the current simulation, we assume that once the ego vehicle reaches the line of sight, there is no ambiguity about whether a pedestrian is present. Hence the expected ambiguity is zero and the epistemic value is scored by the posterior predictive entropy only.

\begin{figure}[!htb]
    \centering
    \includegraphics[width=1\textwidth]{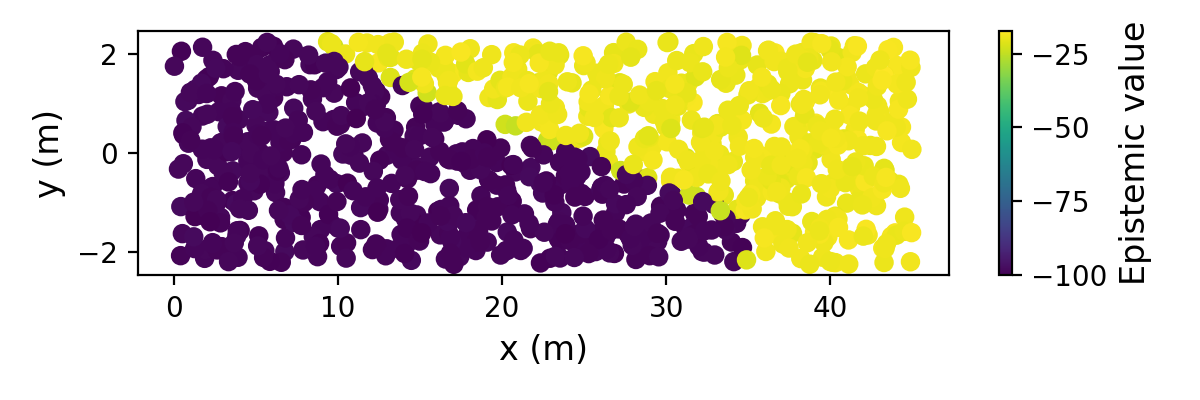}
    \caption{Epistemic affordance for moving left (see the text for explanation).}
    \label{fig:epistemic_map}
\end{figure}

To further illustrate this mechanism, we placed the ego vehicle at different positions on the road map and calculated the epistemic value for each ego position. Figure \ref{fig:epistemic_map} shows a plot of these ego vehicle positions, colored by the corresponding epistemic value (yellow indicates high epistemic value and purple indicates low epistemic value). The figure shows that lateral positions above and beyond the line of sight had about 75 more units of epistemic value than positions below the line of sight. This shows that there is high epistemic value to be gained by moving to the left in the lane, which temporarily trumps the pragmatic value of continuing forward in the center of the lane, leading to the selection of a policy that aims to reach the line of sight sooner. The pattern in Figure \ref{fig:epistemic_map} can be seen as a saliency map representing \textit{epistemic affordances} in terms of future locations from which observations may yield valuable, uncertainty-resolving, information. Once the vehicle passes the line of sight, and uncertainty is resolved, the saliency map will change such that there is no longer a difference in epistemic value along the y dimension, and the driver model will shift back to the lane center, driven by the pragmatic value of maintaining the preferred central lane position.

\begin{figure}[!htb]
    \centering
    \includegraphics[width=1\textwidth]{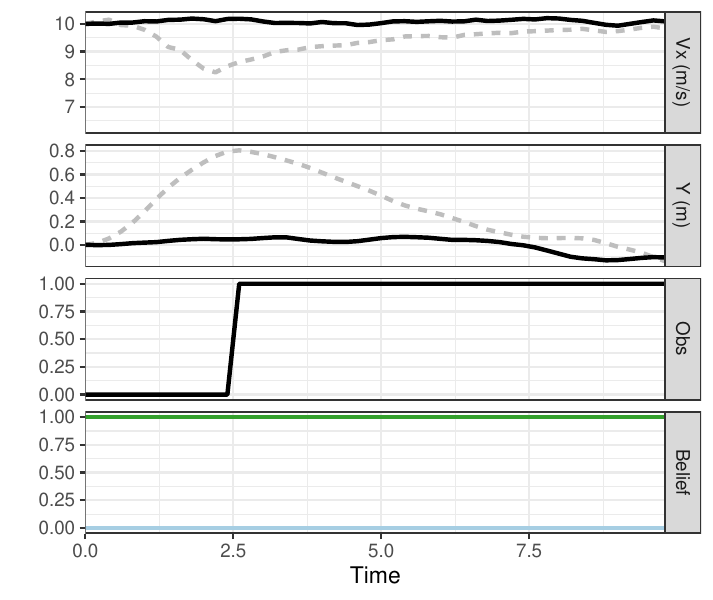}
    \caption{Simulation results for 1d illustrating a scenario where the driver has a strong prior belief that the pedestrian is not present. The vehicle kinematics for this scenario are illustrated in black lines. Grey dashed reference lines illustrate the comparison to the results from Simulation 1c, where uncertainty is present.}
    \label{fig:1d}
\end{figure}

This simulation illustrates one of the key takeaways of this paper: \textit{Since pragmatic and epistemic value are scored in the same currency and optimized under the same objective, selecting policies based on expected free energy minimization allows the driver to find an optimal balance between assertiveness and caution} (conditioned on the drivers beliefs and preferences). This allows the driver to resolve uncertainty “on the fly” simply by moving to a location that provides a better view, an epistemic affordance, which unlocks pragmatic affordances for maintaining efficient progress.
\paragraph{Simulation 1d: Epistemic value depends on the driver’s beliefs}
Figure \ref{fig:1d} shows the results of a simulation where, similar to Simulation 1b, we set the model’s prior belief such that it is (falsely) certain (believes with 100\% probability) that no pedestrian is present. In contrast to Simulation 1c the ego vehicle no longer moves to the left but rather proceeds straight as in 1b. This happens because, from the model’s (false) perspective, there is no uncertainty about the pedestrian and hence no epistemic value to be gained by moving left. Thus, the model’s behavior is, again, driven by pragmatic value only. This is akin to a situation where a human driver is (falsely) certain that no other road users will enter into their path and thus fails to visually scan the road scene (or take other epistemic actions) to the extent that is warranted by the true uncertainty of the situation. 

\subsection*{Scenario 2: Visual time sharing}
In this scenario, we apply a slightly expanded version of the model applied in Scenario 1 to a scenario where the ego vehicle driver is performing a visually demanding secondary task that requires looking away from the road, such as texting on a cellphone. For simplicity, the specifics of the visual task is not modeled, only its effect on glance behavior. That is, we are assuming that the modelled glances away from the road are associated with a visual secondary task that the driver is motivated to perform. Furthermore, we assume that during the off-road glances, no visual information of the road ahead is available to the driver (i.e., ignoring peripheral vision). The driving task involves maintaining longitudinal and lateral vehicle control on a straight road with no other traffic, with a default lane width of 3 m. This set-up is akin to the visual occlusion paradigm \cite{senders1967distribution, pekkanen2017task, pekkanen2018computational}. However, a key difference is that we are here explicitly modeling the motivation to perform the secondary task in terms of a preference (prior distribution) for on-road versus off-road glances.

During off-road glances, the uncertainty about the situation on the road ahead and the vehicle’s position in the lane (i.e.,the two sources of uncertainty identified by \cite{senders1967distribution} discussed above) builds up and the driver needs to intermittently look back to the road to update their beliefs and reset the uncertainty. In the active inference framework, such on-road glances can be seen as actions carrying epistemic value, similar to looking behind the occluding object in Scenario 1 to check for pedestrians. The drive to reduce uncertainty through epistemic actions (on-road glances) is offset by the motivation to look away from the road to perform the secondary task. At the same time, the driver is motivated to keep up speed, maintaining a central lane position and avoid exiting the lane. All Scenario 2 simulations were performed for a segment of 30 seconds. Each scenario was run for 10 times and the time series plots below show one randomly selected run. However, the results from all 10 runs were used to compute summary statistic metrics for comparison with human data.

\subsubsection*{Model specification}
The visual time sharing model builds on the model described under Scenario 1 and, unless otherwise stated, uses the same parameters and values. In this scenario, we replaced the point mass vehicle dynamics model in the previous scenario with a kinematic bicycle model to more accurately capture the vehicle motion constraints \cite{polack2017kinematic}. To generate uncertainty in the ego vehicle’s actual position on the road we introduce a small steering noise (0.001 rad/s) to represent the disturbance caused by uncontrollable environment effects such as uneven road surface or wind gusts. 

We model the state of the driver-vehicle system using the kinematic state of the vehicle in lane coordinates $s^{ego}=[x, y, \theta, \delta, v, a, w]$, two binary variables $Cl$ and $Cr$ for whether the vehicle has crossed the left or right lane boundary, and a binary variable $I$ for the driver’s gaze direction (i.e., off-road or on-road). The lane crossing variables can be seen as representing rumble strips that generate vibrations and sound when crossed, thus conveying to the driver their lane positions even when the driver is looking off-road. The model makes decisions about two types of actions: the kinematic control of the vehicle $a^{control}$, and the gaze action, $a^{I}$, which represents a deterministic transition of the corresponding gaze state. 

To represent the buildup of uncertainty in the model’s belief about the vehicle’s position and heading angle during off-road glances we added steering noise also to the model’s counterfactual steering actions, matching the noise added to the actual steering (0.001 rad/s). Thus, the driver’s generative model matched the generative process. Since the road ahead was always empty, we here only address the second source of uncertainty proposed by Senders et al’s \cite{senders1967distribution} related to vehicle position in the lane. However, the model could be extended to incorporate other sources of uncertainty, for example, related to the behavior of other road users (the first type of uncertainty in \cite{senders1967distribution}).

We define the observations in the same space as the state variables. However, when the driver glances off-road, we assume that the vehicle’s lateral ($y$) and longitudinal ($x$) position and the heading angle ($\theta$) cannot be observed and they are thus set to null values. This means that, due to the noise in the propagation of the beliefs about these states, the uncertainty in the beliefs continue to grow until the driver looks back to the road and the uncertainty is reset. For simplicity, we assume that the belief update when looking back occurs instantaneously during a single time step. That is, we are not modeling the time course of the belief updating process. As a consequence, on-road glances in our simulations typically have a duration of a single time step (200 ms) which is clearly not a realistic representation of visual behavior but sufficient for demonstrating the key principles of the model. This is similar to the occlusion setup where the viewing (visor opening) times also have a fixed duration (250 - 1000 ms in \cite{senders1967distribution}). The state, observation and action variables used in Scenario 2 are listed in Table \ref{tab:Table3}.

\begin{table}[htb]
\caption{State, observation and action variables in the POMDP for Scenario 2}
\label{tab:Table3}
\resizebox{\textwidth}{!}{\begin{tabular}{lcll}
\hline
Variable name & Symbol & Values & Type \\ 
\hline
\multicolumn{4}{l}{States} \\ 
\hline
Gaze & $I$ & off-road (0), on road (1) & Discrete \\
Left bound crossed & $Cl$ & yes (1), no (0) & Discrete \\
Right bound crossed & $Cr$ & yes (1), no (0) & Discrete \\
Ego kinematics & $s^{ego}$ & $x, y, \theta, \delta, v, a, w$ & Continuous \\ 
\hline
\multicolumn{4}{l}{Observations} \\ 
\hline
Gaze observation & $o^{I}$ & off road (0), on road (1) & Discrete \\
Left bound crossed observation & $o^{Cl}$ & yes (1), no (0) & Discrete \\
Right bound crossed observation & $o^{Cr}$ & yes (1), no (0) & Discrete \\
Ego kinematics observation & $o^{ego}$ & $x, y, \theta, \delta, v, a, w$ & Continuous \\ 
\hline
\multicolumn{4}{l}{Action} \\ 
\hline
Ego kinematic control & $a^{control}$ & acceleration (a), steering rate (w) & Continuous \\
Gaze action & $a^{I}$ & off road (0), on road (1) & Discrete \\ 
\hline
\end{tabular}}
\end{table}

The model’s default preferences are generally the same as in Scenario 1, but the lateral acceleration preference is replaced by a preference on steering rate and, as mentioned above, we include two categorical distributions (Cl and Cr) to represent a strong preference for not exiting the lane. Since preferences are defined over observations, during off road glances we do not evaluate the pragmatic value over the variables that cannot be observed when glancing off-road (i.e., vehicle lateral and longitudinal position and the heading angle, as mentioned above). 

We also specify a preference over gaze direction using the log probability of on-road glances based on which we can also compute the log probability of off-road glances to ensure a normalized Bernoulli distribution. The preferences and their default values are defined in Table \ref{tab:Table4}. 

\begin{table}[htb]
\caption{Preferences specifications for Scenario 2}
\label{tab:Table4}
\resizebox{\textwidth}{!}{\begin{tabular}{llc}
\hline
Preference & Specification & Default values \\ \hline
Speed keeping & \begin{tabular}[c]{@{}l@{}}Gaussian distribution \\ centered at the speed limit\end{tabular} & \begin{tabular}[c]{@{}l@{}}$\mu$=10 $m/s$\\ $\sigma$=1 $m/s$\end{tabular} \\
Lane keeping & \begin{tabular}[c]{@{}l@{}}Triangular distribution \\ centered at 0 and bounded at the lane boundaries\end{tabular} & N/A \\
Longitudinal acceleration & \begin{tabular}[c]{@{}l@{}}Gaussian distribution \\ centered at zero\end{tabular} & \begin{tabular}[c]{@{}l@{}}$\mu$=0 $m/s^2$\\ $\sigma$=0.5 $m/s^2$\end{tabular} \\
Steering rate & \begin{tabular}[c]{@{}l@{}}Gaussian distribution \\ centered at zero\end{tabular} & \begin{tabular}[c]{@{}l@{}}$\mu$=0 $rad/s$\\ $\sigma$=0.005 $rad/s$\end{tabular} \\
Gaze preference & \begin{tabular}[c]{@{}l@{}}The log probability of an on-road glance based on a\\ Bernoulli distribution of on-road / off glances.\end{tabular} & -7 \\
Lane exit left / right & \begin{tabular}[c]{@{}l@{}}Categorical distribution \\ representing an absolute preference for not exiting the lane\end{tabular} & N/A \\ 
\hline
\end{tabular}}
\end{table}

\subsubsection*{Simulations}
\paragraph{Simulation 2a: Effects of visual time sharing on vehicle control}
The goal of this first simulation was to establish that our model could generate realistic visual time sharing behavior and reproduce effects of visual time sharing on vehicle control established in the literature. These effects include (1) increased lateral control variability (e.g. \cite{zwahlen1988vision, greenberg2003driver, ostlund2004haste, horrey2006modeling, merat2008effect, engstrom2005effects,mcdonald2020classification}, (2) an increased frequency and magnitude of steering corrections \cite{markkula2006steering, engstrom2005effects, macdonald1980review}
and (3) a reduction in speed (e.g. \cite{merat2008effect, engstrom2005effects, ANTIN1990581}. 

To establish a baseline condition representing vehicle control only (i.e., with no visual time sharing), the preference value for on-road gaze was set to 0 (i.e., the prior probability of on-road glances equals 100\%), resulting in a simulation with no eyes off road glances. We then added a visual secondary task to be performed concurrently with the vehicle control task by lowering the on-road gaze preference from 0 to -7 (hence in effect increasing the preference for looking off road). This can be seen as endowing the model with a motivation to glance off road to perform the visual secondary task. 

\begin{figure}[!htb]
    \centering
    \includegraphics[width=1\textwidth]{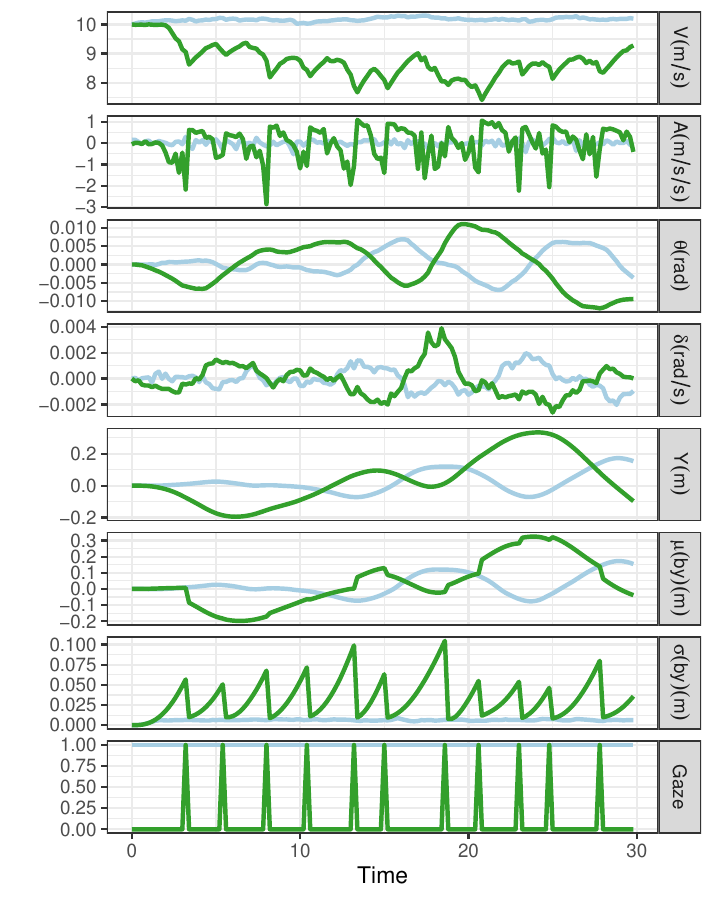}
    \caption{Simulation of baseline (on-road gaze preference = 0; light blue lines) and visual time sharing between the vehicle control and the secondary task (on-road gaze preference = -7; green lines). The charts correspond to the following variables: V = speed, A = longitudinal acceleration, $\theta$ = heading angle, $\delta$ = steering angle, y = lateral position.}
    \label{fig:2a}
\end{figure}

Simulation results for the baseline and visual time sharing conditions are shown in Figure \ref{fig:2a}. As can be seen, the model finds a way to visually timeshare between the driving and the secondary task. When the model is looking off road, the uncertainty in the belief about lateral position (the standard deviation of the belief particles, $\sigma(by)$) increases until it is reset by an on-road glance. By visual inspection of the plot, it can also be observed that the model reduces speed and the variation in lateral vehicle position, and the amount of steering control activity increases during visual time sharing.

\begin{figure}[!htb]
    \centering
    \includegraphics[width=1\textwidth]{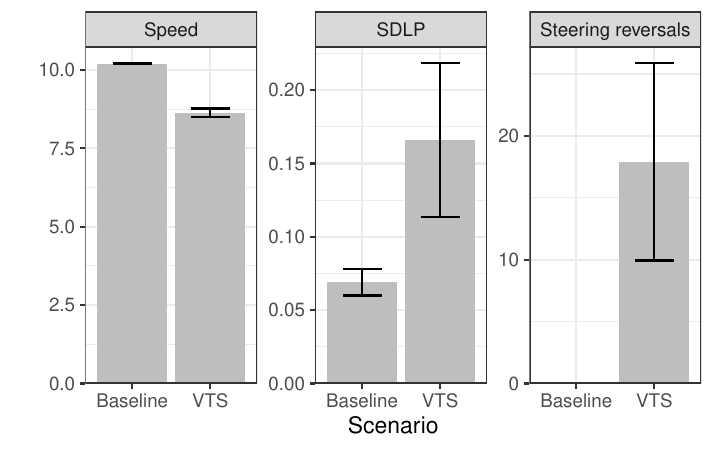}
    \caption{Effects of visual time sharing (VTS; on-road gaze preference = -7) on vehicle control compared to baseline (on-road gaze preference = 0) in the current simulation.}
    \label{fig:2a_control}
\end{figure}

Figure \ref{fig:2a_control} compares summary statistic metrics of speed, lane position and steering reversals. The metrics were mean speed, standard deviation of lane position and the number of large steering reversals, which were computed for all 10 simulation runs, for the baseline and visual time sharing conditions respectively. Steering reversal rate was defined as the number of time steps the front wheel angle exceeds 0.0025 rad/0.14 deg. As shown in the plots, the model reproduces the general effects of visual time sharing on vehicle control in the human data reviewed above, showing a pattern of increased standard deviation of lane position (SDLP), increased rate of large steering reversals and reduced speed. 

The increased lane keeping variability (SDLP) can be generally explained as the result of increased lane drifts that occur during off-road glances driven by the secondary task preference. The increasing number of steering corrections can be explained as corrections to the lane drifts that the driver performs to align the observed lane position with their prior lane keeping preference. Finally, the speed reduction can be understood as a way to settle on an optimal balance between the preference to maximize time on the secondary task and the competing preference to keep the speed close to the speed limit. By reducing speed, the model can “buy” more time for the secondary task (since it takes longer to reach the lane boundary at lower speed), but the model cannot slow down more than mandated by the speed prior preference distribution.

To help explain the underlying visual time sharing mechanism implemented by our expected free energy minimizing model, we visualize the tradeoff between pragmatic and epistemic value in Figure 10. As in Scenario 1, the pragmatic value is determined by the deviation of observations from the modes of the preference prior distributions and the epistemic value is solely driven by the posterior predictive entropy (first term in \ref{eq:epistemic}) since, again, there is no ambiguity in the state-observation mapping (i.e., the second term in \ref{eq:epistemic} equals zero).  

\begin{figure}[!htb]
    \centering
    \includegraphics[width=1\textwidth]{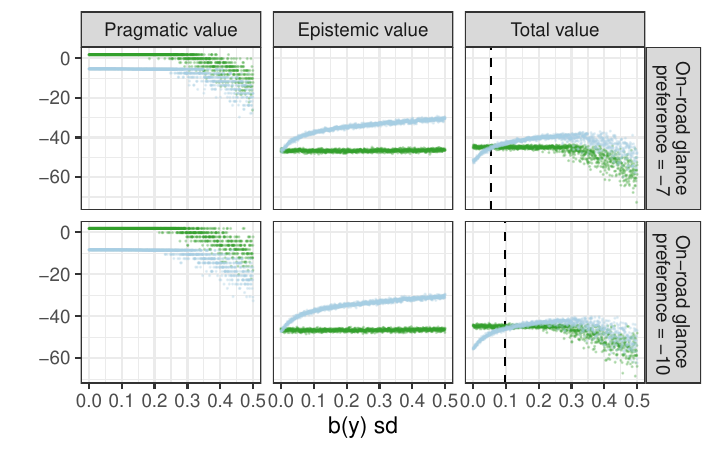}
    \caption{Pragmatic value (left), epistemic value (middle), and total value (right) during  eyes on road (light blue) and eyes off road (green) glances. Top panels: On-road glance preference=-7. Bottom panels: On-road glance preference -10. The dashed reference lines on the total value charts indicate the standard deviation of belief where the total free energy of the off road glance exceeds that of the on road glances.}
    \label{fig:2a_tradeoff}
\end{figure}

Figure \ref{fig:2a_tradeoff} illustrates how the pragmatic, epistemic and total value (negative expected free energy) when gazing off-road (green) vs on-road (blue) vary with increasing standard deviation of artificially generated belief particles representing uncertainty in lateral position. Here, the expected free energy was evaluated for each individual time step (and not for the 4s policy as a whole). The top and bottom panels differ only in the value of the on-road preference. In the top panels we use the default value of -7 and in the bottom panels, we reduced it to -10 (representing a stronger motivation to look off road). In the plots for pragmatic and total value, we truncate the log preference probability of lane exits (a very large negative value) to -100 in order to visualize critical decision points along the vertical axis.

The left panels show that on-road glances generally have lower pragmatic value because they do not fulfill the preference of engaging with the secondary task. The total pragmatic value during on-road glances is higher for the higher on-road glance preference (-7) compared to the lower on-road gaze preference (-10). With increasing dispersion in the belief about lateral position ($b(y)$), the pragmatic values during on-road and off-road glances initially do not change significantly (it is constant for off-road glances and decreases slightly for on-road glances but this is not visible in the figure), but both start decreasing after a cut-off point, reflecting the increased counterfactual risk of exiting the lane. The middle panels show that the epistemic value of an on-road glance increases with increasing uncertainty in the belief about lateral position, which is unaffected by gaze preference (i.e., the top and bottom plots are very similar). Combining the trend in pragmatic and epistemic value, the right panels shows that the total value (negative expected free energy) during an on-road glance exceeds that of an off-road glance at a certain value of $b(y)$ which depends on the gaze preference. At this crossover point (dashed lines in Figure 10), a policy involving a glance back to road will thus be selected. Thus, due to the stronger prior preference for on-road glances, the model in the top panel requires less epistemic value to cross over than the model in the bottom panel with weaker on-road glance preference. Hence, the former model is more prone to look back to the road than the latter when the uncertainty about lateral position increases.

The on-road visual sampling in our model is driven by epistemic value in the same principal way as in the occluded pedestrian scenario in Simulation 1 above. The posterior predictive entropy entropy in (\ref{eq:epistemic}) (which in our model fully determines the epistemic value) of a policy will be highest when the belief is dispersed (high uncertainty) and the model chooses to look back to the road, since this policy is expected to generate a greater variety of possible observations compared to when continuing to look away (and observe nothing) or looking back when the belief is certain (and no new information expected is from looking back). Hence, looking back to the road in this scenario is analogous to reaching the line of sight in the occluded pedestrian scenario.

To summarize, this simulation demonstrates how visual time sharing behavior and its effect on vehicle control emerges from selecting policies that minimize expected free energy. Based on this sole objective, the model strikes a balance between different prior preferences (motivations for the secondary task and vehicle control), and visual sampling of the road ahead is driven by the epistemic value of resolving uncertainty about the current vehicle state (here primarily lateral position). It was also shown that the model explains and reproduces well established effects of visual time sharing on vehicle control in the driver behavior literature. In the following simulations, we further explore how the model accounts for effects of driving task demand and prior preferences on visual time sharing.
\paragraph{Simulation 2b: Effects of driving demand and prior preferences}
As reviewed above, empirical studies has established a strong relationship between demand (difficulty) of the vehicle control task and visual sampling of the forward roadway. In particular, Senders et al. \cite{senders1967distribution} found that increasing driving demand by increasing the set speed led drivers to choose shorter voluntary occlusion intervals (i.e., increased viewing of the road ahead). Similarly, Victor et al. \cite{victor2005sensitivity} found a significantly shorter mean off-road glance duration when driving in curves compared to straight road sections, as well as a trend for more frequent off-road glances in curves.

To explore if our model was able to reproduce and explain these effects, we manipulated driving demand in terms of lane width. It is well established that (all other things being equal) reducing lane width leads to a reduction in speed \cite{yagar1983geometric, fitzpatrick2001design}. 

\begin{figure}[!htb]
    \centering
    \includegraphics[width=1\textwidth]{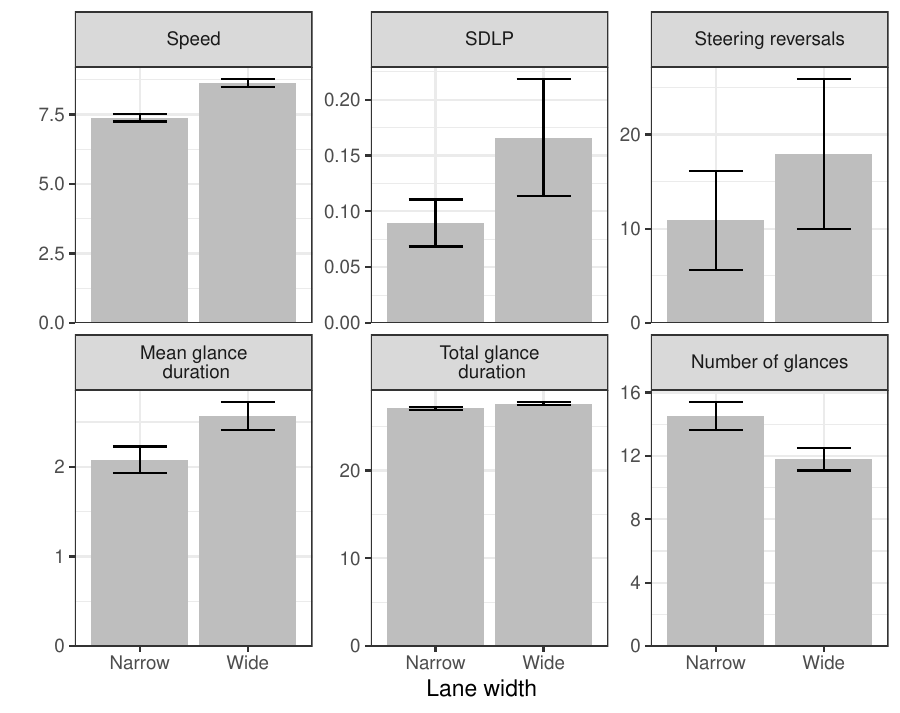}
    \caption{Effects of lane width on vehicle control and glance metrics (wide = 3 m, narrow = 2.5 m).}
    \label{fig:2b_control}
\end{figure}

As shown in Figure \ref{fig:2b_control}, reducing the lane width from 3 to 2.5 m in our simulation leads to a reduction in speed as well as shorter and more frequent off-road glances compared to the wider road, in line with the empirical results from human drivers. This is also accompanied by a reduction in lane keeping variation (SDLP), and less frequent large steering reversals which can be interpreted as a response to the need for tighter lateral control on the narrower road.

In terms of our model, the speed reduction can be explained as an attempt to limit the loss of expected pragmatic value during off-road glances, due to the increased risk of exiting the lane, by slowing down the dispersion of beliefs about future lane position. The reduced duration and increased frequency of off-road glances reflects a need for the model to sample the road more often on the narrower road as driven by epistemic value.

\begin{figure}[!htb]
    \centering
    \includegraphics[width=1\textwidth]{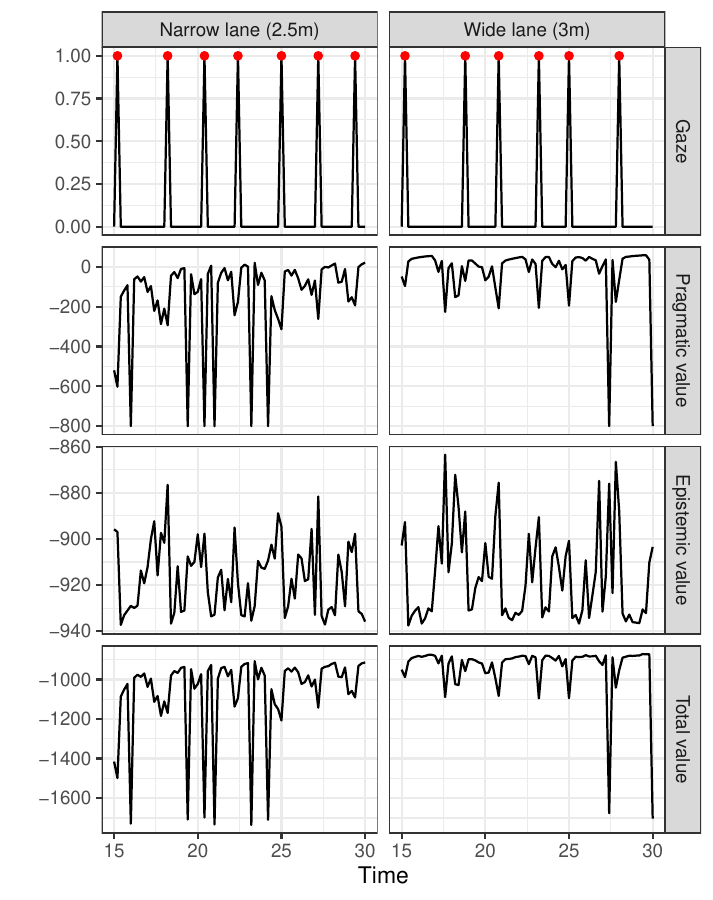}
    \caption{Dynamics of expected free energy minimization during visual time sharing. Large dips in pragmatic value due to lane exits are truncated by setting the lane exit log preference probability to -100, similar to \ref{fig:2a_tradeoff}. The red points in the gaze plot indicate on-road glances.}
    \label{fig:2b_value_decomposition}
\end{figure}

Figure \ref{fig:2b_value_decomposition} further explores the dynamics of the model’s visual behavior and vehicle control and plots the glance behavior along with the overall epistemic and pragmatic value of the optimal (selected) policy at each time step (where the optimal policy is that which minimizes the expected free energy, i.e., maximizes the pragmatic plus epistemic value) during the 4 s planning horizon. The plot zooms in on the last 15 seconds of the simulation segment. It can be seen that, for the wider lane (left panel), each on-road glance is preceded by the selection of policies with increasing epistemic value. Eventually, due to the build-up of uncertainty in lateral position and the corresponding increasing epistemic value of an off-road glance, a policy involving an on-road glance scores the maximum total value and is selected (based on the mechanism illustrated in Figure \ref{fig:2a_tradeoff}). The uncertainty (and the epistemic value) is then reset as the result of the on road glance, leading to the model performing a new off-road glance at the next time step (to maximize pragmatic value in the temporary absence of uncertainty).

For the narrow road (left panel), the pragmatic value is generally lower than on the wider road and is also intermittently reduced during the off-road glances due to the increased proportion of lane exit risk in the selected policies (the large dips in pragmatic value). Thus, less epistemic value is needed to trigger the selection of an on-road glance policy, leading to shorter and more frequent on-road glances (highlighted by the red points in the top charts).

\begin{figure}[!htb]
    \centering
    \includegraphics[width=1\textwidth]{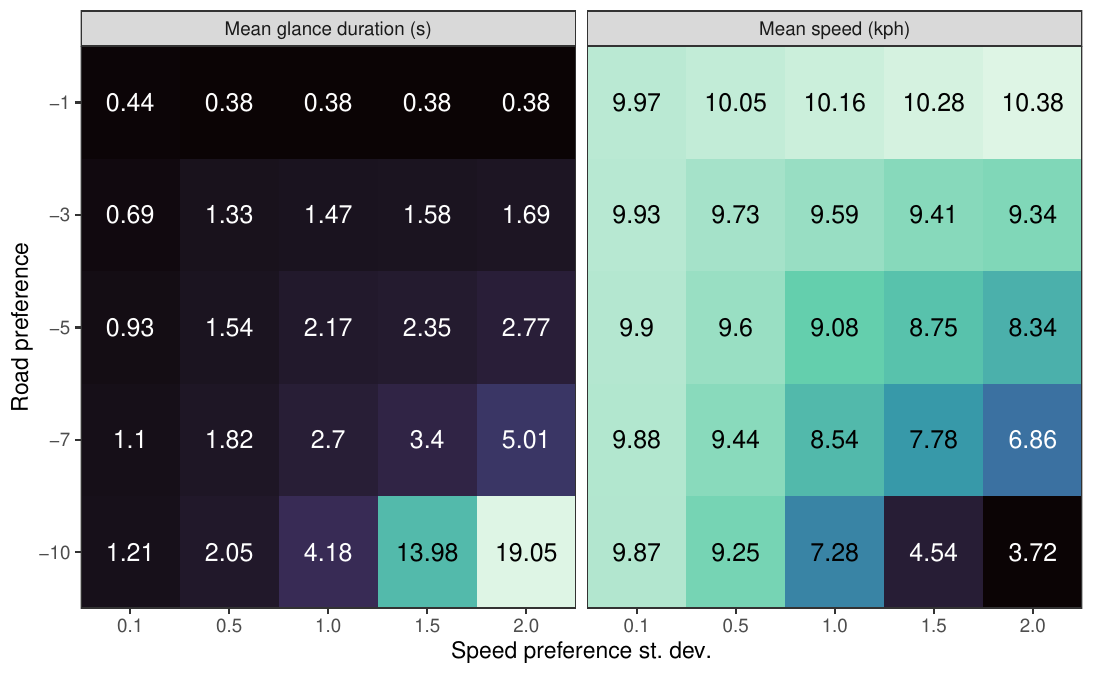}
    \caption{Interaction of on-road glance preference and speed preference on mean speed and mean off-road single glance duration.}
    \label{fig:2b_heatmap}
\end{figure}

Figure \ref{fig:2b_heatmap} explores the tradeoff between speed preference and glance preference by systematically varying on-road glance and speed preference precision on the wide (3 m) lane. Reducing the on-road glance preference (increasing the off-road glance preference) leads to reduced speed but only if the speed preference precision is sufficiently low to allow for speed compensation. This also leads to longer mean off-road single glance durations where, again, the effect is stronger when the speed preference precision is low. Thus, as long as the model is not strongly motivated to keep up the speed, the model adopts a strategy of slowing down to allow for longer off-road glances. By contrast, if the speed keeping cannot be sacrificed due to a high precision of the speed preference prior, the model only looks away briefly before having to look back to the road again. This shows how the visual time sharing strategies adopted by the model depend on the relative motivations for performing the driving and secondary tasks and, more generally, how such strategies emerge naturally from the single mandate of minimizing expected free energy.

\section*{Discussion}
Adaptive driving behavior has been extensively studied for the past sixty years and there exists an influential tradition of (mostly) conceptual models offering explanations for the key underlying mechanisms (e.g., \cite{senders1967distribution, naatanen1976road, summala1988risk, fuller2005towards}). This has been followed by some computational models addressing specific adaptive driving behavior phenomena \cite{kolekar2020human, pekkanen2018computational, johnson2014predicting}. However, so far a unifying computational framework for modeling and explaining the mechanisms underlying these diverse phenomena is lacking. The present paper presented a potential such framework based on active inference \cite{friston2017active, parr2022active} which was demonstrated by a concrete model implementation. 

A common theme in existing models of adaptive driving behavior is that drivers manage the progress versus caution tradeoff by balancing the motivation to achieve goals against the control of uncertainty. Our active-inference-based model implements motivations (preferences) as prior distributions over observations that the driver seeks to realize through behavior (yielding pragmatic value) and the control of uncertainty as information-seeking behavior (yielding epistemic value). Adaptive driving behavior, finding an optimal (given the agent’s beliefs and preferences) balance between progress and caution, then simply emerges by selecting policies and actions that minimize the expected free energy, which can be computed as the (negative) sum of the pragmatic and epistemic value expected in the future.

Pragmatic value can be seen as corresponding to the notion of reward in standard reinforcement learning \cite{sutton2018reinforcement}. However, pragmatic value in the active inference context is conceptually different from the classical notion of (externally imposed) reward in that it is internal to the agent and rests on the idea of self-evidencing and emphasizes the key role of agency: In order to maintain their existence, organisms need to seek evidence for their own generative model of the world \cite{friston2012value}, where the preferred outcomes, defined as prior beliefs over observations, are those expected to be realized through action. This aspect of active inference makes close contact with classical cybernetics models \cite{wallace1960plans} and, in particular, Perceptual Control Theory (PCT), which suggests that behavior can be explained as the control of sensory input signals relative to a reference, or goal, value \cite{powers1973feedback} (see Stephan et al., \cite{stephan2016allostatic} for a discussion about the similarities between PCT and active inference). The self evidencing concept in active inference also makes close contact with ecological psychology \cite{gibson2014ecological} and embodied and enactive approaches to cognitive science (e.g., \cite{clark2013whatever, clark2015surfing, clark2023experience, kiverstein2022problem}).

A key feature in our model is that the strength of the preferences is controlled by the precision (inverse variance) of the preference priors. Thus, when a preference is tuned to high precision, only a narrow range of observations yield high pragmatic value and the model will tend to select policies that do not deviate from this range. This offers a mechanism for protecting a given preference (e.g., keeping the speed near the speed limit) against other conflicting goals (e.g., looking away from the road), and the current model demonstrates how these factors interact in producing different tradeoffs in adaptive behavior (Figure \ref{fig:2b_heatmap}). This can be seen as a potential mechanism underlying \textit{cognitive control}, that is, the ability to adaptively prioritize and focus behavior on the currently most relevant goals and protect against distractions \cite{miller2001integrative, engstrom2017effects}. 

Another key aspect of our model is the resolution of uncertainty in beliefs through epistemic actions. This leads to actions seeking novel, newsworthy, information yielding epistemic value. In the model simulations described above, such epistemic actions included taking a wider turn to get a better view behind an occluding object to resolve uncertainty about conflicts with a potentially hidden pedestrian (Scenario 1), and looking back to the road to resolve uncertainty about one’s position in the lane which had accumulated during the off-road glance (Scenario 2). As shown in this paper, these apparently disparate behaviors both emerge from the same expected free energy minimizing objective.

Epistemic actions may unlock more efficient realization of the preferred observations (pragmatic value), in our case, being able to safely speed up past the occluding object earlier (Scenario 1) and enabling a new off-road glance after the uncertainty about the lane position has been reset (Scenario 2). This seamless interplay between pragmatic and epistemic value can be seen as the core feature of our model (and the active inference framework in general). In active inference, this is enabled by casting the two quantities in the common currency of expected free energy (\ref{eq:efe}). By selecting policies that minimize expected (future) free energy, uncertainty can be resolved “on the fly” as an integral part of the generated behavior. As we demonstrate in this paper, a single action often carries both pragmatic and epistemic value, for example, in Simulation 1a where the ego vehicle was moving forward towards the goal (yielding pragmatic value) which also brought about a better view of the area behind the occluding object (epistemic value). 

Epistemic value has also been explored in machine learning where it is typically referred to as artificial curiosity and intrinsic motivation \cite{schmidhuber1991curious, sun2011planning, hester2012intrinsically}. In this context, the purpose of curiosity and intrinsic motivation  is generally to facilitate model learning by promoting a wider exploration of the state space. By contrast, the focus of the present paper is on the role of epistemic action in resolving uncertainty during inference. However, active inference also extends to learning, which can also be cast in terms of expected free energy minimization and facilitated by epistemic action \cite{friston2015active}. While classical machine learning approaches to exploration often involve ad-hoc features and/or separate mechanisms for exploitation and exploration, active inference has the advantage of offering a more principled approach for both inference and learning solely based on expected free minimization.

Minimizing expected free energy “automatically” yields an optimal combination of epistemic and pragmatic action, but the ensuing behavior is only optimal given the agent’s subjective preferences and beliefs. This allows for conceptualizing and modeling situations where a driver’s understanding of the situation is incorrect, that is, the driver’s generative model does not match the actual state of affairs (the generative process) but the behavior is still optimal given the agent's subjective beliefs and preferences. In this paper, we demonstrated how an overly certain (high precision) belief may expose the driver to an increased risk for collision. In our Simulations 1b and 1d, the model falsely believed with full certainty that no pedestrian could be hidden behind the occluding object, in which case it did not slow down or moved left to get a better view. Based on the model’s own beliefs and preferences, this represented the optimal behavior in this situation but the model’s beliefs were not well calibrated to the situation, representing the general phenomenon of optimal inference with sub optimal models studied in computational psychiatry \cite{schwartenbeck2016inference}. It is widely believed that failures to properly adapt behavior to the traffic situation due to false beliefs is a leading factor behind road crashes \cite{summala1996accident, summala2007towards}.

In Scenario 2, we demonstrated that our model was able to reproduce and explain the underlying mechanisms behind key human behavioral patterns in the literature on visual sampling and visual time sharing in driving. Specifically, the model explains how visual time sharing emerges as a result of pragmatic value motivating off road glances to perform a secondary task, counteracted by the epistemic value of resolving uncertainty about one’s position in the lane, all governed by selecting policies that minimize expected free energy. At the high level, our model shares many of the key concepts behind the existing computational visual sampling models by Senders et al. \cite{senders1967distribution}, Pekkanen et al. \cite{pekkanen2018computational} and Johnson et al., \cite{johnson2014predicting}), in particular regarding the explicit modeling of uncertainty and the key role of visual sampling in resolving uncertainty. However, whereas the existing models are specifically applicable to visual sampling, our model is, in principle, applicable to any form of adaptive driving behavior. While the current relatively simple scenarios were chosen to illustrate the key principles of the model, we believe that our model, thanks to its generality, can in principle be applied to any traffic scenario. Even though the specific implementation details may differ, the current framework suggests that all that is needed to model certain behavior is to define the preference distribution, belief distribution, the observation and action variables (as probability distributions), and the generative model for how they evolve. Adaptive driving behavior will then fall out “automatically” by selecting policies that minimize expected free energy. 

While the current implementation of the model is based on active inference principles (specifically expected free energy minimization), it does not represent a “pure” implementation of the type typically found in the active inference literature (e.g., \cite{parr2022active}). Rather the model makes use of standard engineering methods such as particle filtering (for representing beliefs about future states) and the cross entropy method for policy selection. A related approach is presented in Fountas et al. \cite{fountas2020deep} which used Monte Carlo Tree Search to generate policies evaluated through expected free energy minimization. This yields a modular architecture where the different components can be substituted for other models and methods. For example, in order to represent more sophisticated probabilistic beliefs (e.g., about the future behavior of other road users), the particle filter in the current model could be replaced by a more advanced machine-learned behavior prediction model (e.g., \cite{chai2019multipath}), while still retaining the key principles of active inference discussed above. 
 
Whereas, in the current model, the parameter values were set by hand, the model naturally lends itself to learning the parameters from data (see \cite{wei2022modeling, wei2023active} for our earlier work in this direction). The design space for incorporating techniques from contemporary generative AI in active inference models is large and the exploration of these possibilities has only begun (see e.g. \cite{tschantz2020scaling, mazzaglia2022free, lanillos2021active, fountas2020deep, friston2022designing}). However, regardless of the implementation, behavioral models developed based on the active inference principles outlined above are fundamentally explainable and interpretable which is one of their key potential advantages compared to existing black-box approaches for agent modeling \cite{albarracin2023designing}.  

The modeling framework outlined in this paper has many potential applications in road traffic research beyond driver agent modeling. For example, it can be used as the basis for defining and modeling human road user failure modes behind road crashes, and for designing effective countermeasures. As discussed above, one such mechanism is how too high precision (false certainty) in one’s belief in how a situation will play out leads to overly assertive behavior (e.g., not accounting for potential occlusions, and increased crash risk). While such crash causation mechanisms have been previously outlined conceptually \cite{engstrom2018great}, the current model offers a precise computational formulation of these principles. The model can also be used to derive reference trajectories for the evaluation of autonomous vehicles which are optimal given preferences that reflect the societal norms in the community where the AI is deployed and reasonably foreseeable (assumptions (beliefs) about the behavior of other road users (as currently formalized in \cite{rightieee}).

The model presented in this paper has a number of specific limitations related to the simplifying assumptions of the current implementation described above (e.g., assuming only a single pedestrian in Scenario 1, ignoring peripheral vision in Scenario 2 etc.). In general we don’t think these simplifying assumptions imply any fundamental limitations for the proposed behavior modeling framework. However, there are certain aspects of the current model set-up that warrants further consideration. For example, the current model assumes that some variables (vehicle x and y position and heading) are not observable when looking off-road, and that this also applies during the counterfactual evaluation of policies. During these counterfactual off-road glances, the pragmatic value related to vehicle x and y position and heading cannot be evaluated. This is a consequence of pragmatic value being defined over observations and necessitated the addition of the lane exit variable which is assumed to be always observable. Clearly, this is somewhat arbitrary and an alternative approach would be to define pragmatic value during counterfactual reasoning over states instead of observations, in which case only the lane position variable would be needed (and the lane exit variable would be redundant). Such alternative model variants can be explored in further work.
 
More broadly, there are many ways in which the current model can be further developed. First, as already mentioned, using more advanced generative models for representing beliefs and learning parameters from data will likely be needed to scale up to more realistic scenarios. Second, it would be interesting to explore further the extent to which the model can generalize across different types of traffic scenarios. Of particular interest is to model collision avoidance behaviors in critical traffic conflicts. Such behaviors are difficult to learn solely from data (due to the sparsity of such long-tail events in human data), while the present model could be better suited for such scenarios thanks to the possibility to explicitly specify the behavioral mechanisms. This would require a detailed model of the belief updating process to model response timing, a feature which is lacking in the current model (see \cite{engstrom2022modeling}). Another very interesting avenue of future model development is in the context of road user interactions, for example at a crosswalk \cite{markkula2023explaining}. In such scenarios, active inference models can represent the situation understanding shared between interacting road users in terms of shared generative models (see \cite{friston2015active, friston2022designing}) and, conversely, failing interactions can be understood in terms of mismatching generative models.

Another interesting question not addressed in the present paper concerns how to model the difference between explicit planning and more automatized habitual behavioral selection. The current model is based on sampling and rolling out a large number of policies (plans) where the optimal plan is identified based on expected free energy minimization. However, human action selection in everyday situations is at least partly governed by habits. In skilled performance, humans do not evaluate all possible policies but rather learn to “see” the best policy given the information currently available (e.g., in terms of Gibsonian affordances; \cite{gibson2014ecological}). This is particularly true in the context of driving which is often to a large extent automatized for experienced drivers. In current ML-based active inference models, such habit formation has been modeled in terms of  “habit networks” which are able to learn successful policies through amortization (e.g., \cite{mazzaglia2022free}) and this is an interesting topic for future development of the present framework.

%\section*{Conclusion}

%\section*{Supporting information}

%\section*{Acknowledgments}

\nolinenumbers

\bibliography{ref}  % .bib
\end{document}